\documentclass[11pt]{article}
\usepackage[T1]{fontenc}
\usepackage{lmodern}
\usepackage[margin=1in]{geometry}
\usepackage{amsmath,amssymb,amsthm,mathtools}
\usepackage{microtype}
\usepackage{aliascnt}
\usepackage{booktabs,array}
\usepackage{titlesec}
\usepackage{xcolor}
\usepackage[colorlinks=true,linkcolor=blue!65!black,citecolor=blue!65!black,urlcolor=blue!65!black]{hyperref}
\usepackage[capitalize,noabbrev,nameinlink]{cleveref}
\usepackage{makecell}
\hypersetup{pdftitle={Observable-Specific Stability of Diffusion Sampling: A Transport and Inverse-Spectral Perspective},pdfauthor={}}
\numberwithin{equation}{section}
\newtheorem{theorem}{Theorem}[section]
\newaliascnt{corollary}{theorem}
\newtheorem{corollary}[corollary]{Corollary}
\aliascntresetthe{corollary}
\crefname{corollary}{corollary}{corollaries}
\Crefname{corollary}{Corollary}{Corollaries}
\newaliascnt{proposition}{theorem}
\newtheorem{proposition}[proposition]{Proposition}
\aliascntresetthe{proposition}
\crefname{proposition}{proposition}{propositions}
\Crefname{proposition}{Proposition}{Propositions}

\newaliascnt{claim}{theorem}
\newtheorem{claim}[claim]{Claim}
\aliascntresetthe{claim}
\crefname{claim}{claim}{claims}
\Crefname{claim}{Claim}{Claims}

\newaliascnt{lemma}{theorem}
\newtheorem{lemma}[lemma]{Lemma}
\aliascntresetthe{lemma}
\crefname{lemma}{lemma}{lemmas}
\Crefname{lemma}{Lemma}{Lemmas}
\theoremstyle{definition}
\newaliascnt{definition}{theorem}
\newtheorem{definition}[definition]{Definition}
\aliascntresetthe{definition}
\crefname{definition}{definition}{definitions}
\Crefname{definition}{Definition}{Definitions}
\newaliascnt{assumption}{theorem}
\newtheorem{assumption}[assumption]{Assumption}
\aliascntresetthe{assumption}
\crefname{assumption}{assumption}{assumptions}
\Crefname{assumption}{Assumption}{Assumptions}
\theoremstyle{remark}
\newaliascnt{remark}{theorem}
\newtheorem{remark}[remark]{Remark}
\aliascntresetthe{remark}
\crefname{remark}{remark}{remarks}
\Crefname{remark}{Remark}{Remarks}
\DeclareMathOperator{\Var}{Var}

\DeclareMathOperator{\Law}{Law}
\DeclareMathOperator{\TV}{TV}

\newcommand{\RR}{\mathbb{R}}
\newcommand{\EE}{\mathbb{E}}
\newcommand{\grad}{\nabla}
\newcommand{\norm}[1]{\left\| #1 \right\|}
\newcommand{\phat}{\widehat p}
\newcommand{\normal}{\mathcal N}
\newcommand{\Zrev}{\overleftarrow{Z}}
\DeclareMathOperator{\poly}{poly}
\newcommand{\shat}{\widehat s}
\newcommand{\epsscore}{\varepsilon_{\mathsf{score}}}
\newcommand{\spectralineq}[1]{$#1$-conditioned Poincar\'e inequality}
\newcommand{\TT}{\mathsf T}
\DeclareMathOperator{\diag}{diag}
\newcommand{\eps}{\varepsilon}
\newcommand{\ScoreErr}[2]{\mathsf{ScoreError}_{#1}\!\left(#2\right)}
\newcommand{\NN}{\mathbb{N}}
\newcommand{\Dchi}[2]{\chi^2(#1\Vert #2)}
\newcommand{\Dkl}[2]{\mathsf{KL}(#1\Vert #2)}
\newcommand{\dd}{\mathrm{d}}

\newcommand{\ol}{\overline}
\newcommand{\Vstar}{V^\star}
\usepackage{bbm}
\newcommand{\wh}{\widehat}
\newcommand{\cX}{\mathcal{X}}
\newcommand{\laplace}{\triangle}
\newcommand{\cA}{\mathcal{A}}
\newcommand{\tstar}{t^\star}
\newcommand{\tcrit}{t_{\mathsf{crit}}}
\usepackage{xcolor}

\newcommand{\dhruv}[1]{\textcolor{red}{[DR: #1]}}

\newcommand{\CC}{\mathcal{C}}
\newcommand{\WW}{\mathsf{W}}
\newcommand{\indic}{\mathbbm{1}}

\newcounter{parmarker}

\title{Error accumulation in reverse diffusion with an imperfect score: beyond path-space analyses}

\author{}
\date{}

\begin{document}
\maketitle

\vspace{-3em}
\begin{center}
\setlength{\tabcolsep}{18pt}
\begin{tabular}{cccc}
\makecell{George Giapitzakis \\ \small{CMU}}
&
\makecell{Ankur Moitra \\ \small{MIT}}
&
\makecell{Andrej Risteski \\ \small{CMU}}
&
\makecell{Dhruv Rohatgi \\ \small{UC Berkeley}}
\end{tabular}
\end{center}
\vspace{1em}

\begin{abstract}
Diffusion modeling is a popular method of sampling from a complex, high-dimensional distribution $p$ using learned approximations of its (noised) \emph{score functions}. Generating a single sample requires simulating the reverse diffusion process, which requires many queries to the learned scores. Thus, a priori, the learning errors can easily compound. Indeed, in existing theoretical analyses, the sampling error (as measured in total variation distance to $p$) accumulates polynomially with the horizon of the stochastic process. This is a consequence of bounding the error of the output law by the error of the law of the entire reverse diffusion \emph{path}.

In this paper, we demonstrate that by circumventing the ``path-space'' analysis, the error accumulation of the diffusion sampler can be mitigated when $p$ satisfies a Poincar\'e inequality. Moreover, we achieve this result as a special case of a broader framework that applies to \emph{any} distribution $p$, but relaxes the error metric from total variation distance to a weaker metric capturing all sample statistics that do not ``witness'' the multimodality of $p$.

We then zoom out from analyzing the diffusion sampler to the information-theoretic question: given a learned score with some error bound, how accurately can we identify $p$? We show that, for any given bound on the Poincar\'e constant of $p$, the diffusion sampler already nearly achieves optimal total variation error.

\end{abstract}


\section{Introduction}\label{sec:introduction}

One of the main design principles in the dominant generative models---whether they are autoregressive language models \cite{vaswani2017attention,radford2019language,bai2023qwen} or diffusion/flow-based generative models \cite{ho2020denoising,dhariwal2021diffusion,esser2024scaling,zhao2026qwen}---is to \emph{generate through a sequential process}, learning the prediction rule used at each step via \emph{supervised learning}---with supervision constructed automatically from the data itself. 
For example, autoregressive language models generate text by sequentially \emph{generating the next token}, given the previous tokens; next-token prediction is a supervised learning task with abundant existing training data \cite{vaswani2017attention}. For denoising diffusion models, the sequential structure is less apparent from the data, but nevertheless natural: images are generated by starting from white noise, and then iteratively applying a denoising operator to remove a small (in the continuous limit, infinitesimal) amount of noise \cite{ho2020denoising}. Again, each denoising step can be framed as a supervised learning task where training data can be automatically generated from real images by adding Gaussian noise.

The sequential structure allows richer supervision, and thus makes learning easier. 
However, because generation involves chaining multiple learned steps, the errors of the individual steps can be \emph{amplified}. This phenomenon has been empirically observed or hypothesized in numerous settings, including both language modeling \cite{holtzman2019curious} and diffusion modeling \cite{pmlr-v202-ning23a,li2024error}, but the fundamental sources remain poorly understood. This paper investigates error amplification specifically in the case of \emph{diffusion models}. 

\subsection{Background and main questions}
(Denoising) diffusion models \cite{sohl2015deep,song2020score,ho2020denoising} generate samples by approximately simulating a stochastic differential equation (SDE) called the \emph{reverse SDE}, which reverses the \emph{forward SDE} which gradually adds Gaussian noise to a clean image and converges to a standard Gaussian. Thus, the reverse SDE starts with a sample from a standard Gaussian, and finishes with a sample from the image distribution. The reverse SDE depends on the \emph{score function} of every marginal distribution along the diffusion path (see \Cref{sec:setting} for formal definitions)---which is what the training loss is designed to fit. 

The typical way to solve the reverse SDE---and thus produce a sample---is to use a \emph{numerical discretization} schema. Characterizing how close these samples are to the target distribution requires characterizing three sources of error:

\begin{enumerate} 
    \item \textbf{Initialization error:} in practice, the reverse SDE is initialized with Gaussian noise at some finite time $T$, which we term the \emph{diffusion horizon}; this initialization is not quite correct (except in the limit $T \to \infty$).  
    \item \textbf{Discretization error:} the numerical discretization schema will not perfectly track the continuous-time stochastic process, leading to errors throughout generation. 

    \item\textbf{Score error:} the score functions are learned from data, so they will not exactly match the true scores. This is another source of errors throughout generation.
\end{enumerate}

Classical theoretical work on diffusion models \cite{chen2023sampling, lee2022convergence} characterized these sources of error, to obtain \emph{total variation distance} bounds on the output law $\phat$ of the diffusion sampler. They showed 
that if $p$ is the $d$-dimensional data distribution, then under mild regularity conditions, the output law $\phat$ satisfies the following bound, with one term from each of the above sources:
\begin{equation}\TV(p,\phat) \lesssim e^{-T}\sqrt{\Dkl{p}{\normal(0,I_d)}} + \frac{\poly(d, T)}{N} + \epsscore\sqrt{T},\label{eq:standard-girsanov-analysis}\end{equation}
where $T$ is the diffusion horizon, $N$ is the number of discretization steps, and $\epsscore^2$ bounds the $L^2$ error in the learned score. Notably, the score error term \emph{grows with $T$}. To balance the initialization error and score error, one would choose $T \asymp \log(\Dkl{p}{\normal(0, I_d)}/ \epsscore)$. Even for ``nice'' distributions like $p := \normal(0, 2I_d)$, the KL divergence can be as large as $\poly(d)$, thus requiring $T \asymp \log(d/\epsscore)$. 

The $\sqrt{T}$ factor in the score error term arises from 
bounding $\TV(p,\phat)$---the distance between the terminal marginal laws of the true and learned reverse SDEs---by the distance between the respective \emph{path laws}; this distance can in turn be controlled by Pinsker's inequality and Girsanov's theorem \cite{girsanov1960transforming}---a change-of-measure formula that enables writing the KL divergence between the path laws of two SDEs in terms of their $L^2$ difference in drifts. This chain of bounds can be pessimistic: in particular, discrepancies between entire path laws may be substantially larger than discrepancies between their terminal marginals, since early errors may be attenuated by subsequent dynamics.

Nevertheless, subsequent theory on error bounds for diffusion models has almost uniformly written off the $\epsscore\sqrt{T}$ term as unavoidable, and has instead focused on the discretization error---e.g. improving its dimension dependence under the manifold assumption \cite{debortoli2022convergence,li2024adapting,huang2026denoising}, or improving the query complexity of reaching $\epsscore\sqrt{T}$ error \cite{gatmiry2026high,chen2026highaccuracy}. 

In this paper, we take a complementary perspective: putting aside discretization issues (which can be mitigated with more computation), when can we \emph{avoid} the $\epsscore\sqrt{T}$ error incurred by the path-space analysis---in other words, when do score errors actually accumulate? There are two natural versions of this question:

\vspace{1em}
\noindent\textbf{Question I.} Are there \emph{structural assumptions} on the data distribution $p$ under which the score error is amplified by less than $\sqrt{T}$, when  bounding $\TV(p,\phat)$?

\vspace{1em}
\noindent\textbf{Question II.} For general data distributions, are there \emph{weaker error metrics} (than total variation distance) in which this amplification can be reduced?

\vspace{1em}
We will tackle these questions simultaneously by formulating an even broader question: given a distribution $p$, what measurable ``properties'' of $p$ will the diffusion sampler approximately preserve? 

To state this broader question concretely, recall that the total variation distance has a dual formulation as a supremum over bounded \emph{test statistics}:
\[\TV(p, \phat) := \sup\left\{|\EE_p[f] - \EE_{\phat}[f]| \middle| f\colon \RR^d \to [0,1]\right\}.\]
Naturally, some test statistics will witness larger discrepancies than others, and the structure of $p$ may influence which is which. The standard path-space analysis does not distinguish how sensitive different test statistics $f$ are to score errors. In contrast, our analysis will bound the discrepancy $|\EE_p[f] - \EE_{\phat}[f]|$ independently for each $f$, thereby retaining dependence on $f$. This perspective is also motivated by empirical observations that diffusion models preserve some properties of the data distribution more accurately than others \cite{perera2023analyzing,seshadri2024bias,aithal2024understanding}.

We introduce a \emph{geometric condition} on a test statistic $f$ relative to the distribution $p$ that allows for improved control of error propagation. A global Poincar\'e inequality implies this condition uniformly over \emph{all bounded statistics}, yielding an improved total variation guarantee. For general distributions, this condition yields improved bounds for \emph{restricted families of statistics}\footnote{Equivalently, in integral probability metrics weaker than total variation.}, even when the distribution lacks such global geometric control. Thus, the two questions lead to a common theory of how the geometry of $p$ governs score-error accumulation for different statistics.

\subsection{Our contributions}\label{sec:contributions}

We use the term ``test statistic'' to refer to a functional $f:\RR^d \to \RR$ (satisfying mild regularity properties; see \Cref{sec:setting}). Our first main result bounds the error accumulation of the diffusion sampler, as measured by a given bounded test statistic $f$, in terms of a geometric property that we introduce and call the \spectralineq{f}. We first (informally) state the result, then explain the definition of this inequality, as well as concrete applications of the result. 

In this overview, we let $\phat$ denote the output law of the continuous learned reverse SDE, run for time $T$ with Gaussian initialization. As above, we let $\epsscore^2$ denote an upper bound on the $L^2$ errors of the score functions. We defer formal definitions to \Cref{sec:setting}.

\begin{remark}[On discretization]\label{remark:discretization}
We analyze the continuous SDE in order to clarify the focus of the paper; as previously mentioned, there is by now extensive work on discretization in diffusion models, and we believe our results should extend straightforwardly to any standard discretization schema (with an additional error term scaling inverse-polynomially with the number of steps, as in \eqref{eq:standard-girsanov-analysis}). 
\end{remark}

\begin{theorem}[Informal; see \Cref{thm:observable-main}]\label{thm:observable-intro}
Let $f$ be a $[0,1]$-valued test statistic. Then $\phat$ satisfies
\begin{align}
 |\mathbb E_{\widehat p}f-\mathbb E_p f|
 &\lesssim e^{-T}\sqrt{\Dkl{p}{\normal(0,I_d)}}+\epsscore\sqrt{1+\log(1+C_P(p;f))}
       +\epsscore^2T.
 \label{eq:intro-whole-observable-bound}
\end{align}
where $C_P(p;f)$ is the constant in the \spectralineq{f} for $p$ (\Cref{def:spectral-ineq}).
\end{theorem}

For small $\epsscore$, the second term in \eqref{eq:intro-whole-observable-bound} dominates the third; thus \eqref{eq:intro-whole-observable-bound} represents more benign error accumulation compared to the standard path-space bound \eqref{eq:standard-girsanov-analysis}, albeit under an additional condition---the \spectralineq{f}.
This condition is most easily explained geometrically. Writing $g:=f-\EE_p[f]$, consider the perturbed distribution
\[
p^{\delta,g}:=(1+\delta g)p.
\]
The \spectralineq{f} bounds the transport cost of this infinitesimal reweighting, with the variance normalizing the size of the perturbation:
\[
\TT_p(g)
:=
\lim_{\delta\to0}
\frac{\WW_2^2(p,p^{\delta,g})}{\delta^2}
\le C_P(p;f)\Var_p(f).
\]
For intuition, suppose $p=\sum_i w_iq_i$ is a mixture of well-separated components, each satisfying a Poincar\'e inequality with constant at most $K$. If $\EE_{q_i}[f]$ is the same for every component, the reweighting preserves each component's total mass and can be realized by transport within components: indeed, we can show that $C_P(p;f)\le K$ (\Cref{prop:mixture-flux}). By contrast, if these expectations differ, the reweighting changes the relative component weights and may require costly transport between modes. The condition therefore distinguishes statistics according to how their associated perturbations interact with the geometry of $p$. Formally, this condition is provided as \Cref{def:spectral-ineq}.

\paragraph{Applications of \Cref{thm:observable-intro}.} The simplest application is when $p$ itself satisfies the standard Poincar\'e inequality. In this case, the \spectralineq{f} holds for every $f$, with the same constant. Thus, an immediate corollary of \Cref{thm:observable-intro} is a ``beyond path-space'' total variation bound for the learned reverse SDE, whenever $p$ satisfies the Poincar\'e inequality:

\begin{corollary}[Informal; see \Cref{cor:poincare-main}]\label{cor:poincare-intro}
It holds that
\begin{align}
 \TV(p,\phat)
 &\lesssim e^{-T}\sqrt{\Dkl{p}{\normal(0,I_d)}}+\epsscore\sqrt{1+\log(1+C_P(p))}
       +\epsscore^2T.
\end{align}
where $C_P(p)$ is the constant in the Poincar\'e inequality for $p$ (\Cref{def:poincare}).
\end{corollary}

Interestingly, \Cref{cor:poincare-intro} can be combined with the standard path-space analysis to tighten \eqref{eq:standard-girsanov-analysis} in the setting where $p$ is a generic distribution supported in the centered ball of radius $R$. The idea is that even though $p$ itself may not satisfy a Poincar\'e inequality, the forward diffusion process will quickly ``smooth'' $p$ into a distribution that does. See \Cref{cor:bounded-support} and \Cref{remark:bounded-support} for details. 

As an additional application of \Cref{thm:observable-intro}, consider again the setting where $p = \sum w_i q_i$ is a mixture of components that individually satisfy the Poincar\'e inequality with constant at most $K$. As mentioned previously, we have $C_P(p;f) \leq K$ for any test statistic $f$ for which $\EE_{q_i}[f]$ is the same for every component.
As a result, by \Cref{thm:observable-intro}, such statistics are learned accurately (\Cref{cor:balanced-main}), even though other statistics that discriminate between components may incur error accumulation. This corroborates empirical evidence suggesting that diffusion models may err in estimating the relative weights of different classes, despite achieving high perceptual quality \cite{perera2023analyzing, seshadri2024bias}.\footnote{The work of \cite{seshadri2024bias} on biases in text-to-image models suggests that much of the observed bias amplification from training to test set can be explained by a change in prompt distribution, but not all.}

\paragraph{High-level intuition for proofs.} We briefly explain why results of the above form are plausible, focusing on \Cref{cor:poincare-intro} and deferring a full overview to \Cref{sec:overview}. The key intuition why we might expect the score error to amplify by only $\sqrt{\log(C_P(p))}$, rather than $\sqrt{T}$, is that the forward diffusion attenuates the Poincar\'e constant to $e^{-2t} C_P(p) + 1$ in $t$ time; thus, the marginal distributions $(p_t)_t$ are fast-mixing outside the interval $[0, \log(C_P(p))]$, and so one may hope that the final output of the learned reverse diffusion is only sensitive to errors made within this interval.

This intuition can be formalized, for any test statistic $f$, via the \emph{posterior statistics} $f_t(x) := \EE[f(Z_0) \mid{} Z_t = x]$, where $(Z_t)_t$ is the forward diffusion process. Indeed, by an adjoint estimate (\Cref{lem:whole-comparison}), the final error under $f$ has leading-order term bounded by
\begin{equation}
2\epsscore \int_0^T\sqrt{\EE_{p_t}\norm{\grad f_t}^2}\,\dd t, \label{eq:integratedbd}\end{equation}
where the integrand captures the sensitivity to local errors made at time $t$. The central idea now is that the natural ``clock'' for the integral \eqref{eq:integratedbd} is not the diffusion time $t$, but rather the \emph{leftover variance} $\Var_{p_t}(f_t)$, because of the identity $\frac{\dd}{\dd t} \Var_{p_t}(f_t) = -2\EE_{p_t}\norm{\grad f_t}^2$---i.e. the diffusion process necessarily depletes variance when $f_t$ has large gradients (\Cref{lem:posterior-variance}). To obtain a horizon-independent bound on \eqref{eq:integratedbd}, it suffices to argue that $\Var_{p_t}(f_t)$ decays exponentially fast, which follows from the Poincar\'e inequality:
\[\frac{\dd}{\dd t} \Var_{p_t}(f_t) = -2\EE_{p_t}\norm{\grad f_t}^2 \leq -\frac{2}{C_P(p_t)} \Var_{p_t}(f_t).\]
This final step is analogous to the standard argument that Langevin dynamics exponentially contracts $\chi^2$-divergence \cite{Chewi26Book}---except we incur only logarithmic dependence on $C_P(p)$, rather than linear, because $C_P(p_t)$ itself decays exponentially.

This argument readily generalizes to prove \Cref{thm:observable-intro}, since the Poincar\'e inequality for $p_t$ is only ever applied to the statistic $f_t$. The missing piece is to argue that the \spectralineq{f} \emph{propagates} to the posterior statistics in the same manner that the Poincar\'e inequality for $p$ propagates to $p_t$. See \Cref{sec:overview} for a detailed overview.

\paragraph{Zooming out: information-theoretic limits.} The diffusion sampler is just one possible method of sampling using an estimated score (albeit one that has received the most attention both in theory and practice). It's reasonable to ask whether there are better methods that achieve even less error.

We show that information-theoretically, when $p$ satisfies a standard Poincar\'e inequality, it is possible to mitigate the error accumulation even further, and remove the initialization error and higher-order term $\epsscore^2 T$ from \Cref{cor:poincare-intro}:

\begin{proposition}[Informal; see \Cref{prop:it-upper-main}]\label{prop:it-upper-intro} 
Given the estimated score functions, it is possible to identify a distribution $q$ satisfying
\[\TV(p, q) \lesssim \epsscore \sqrt{1+\log(1+C_P(p))}.\]
\end{proposition}

Unfortunately, the algorithm to do so is computationally inefficient: it requires projecting the estimated score functions to the nearest family of vector fields $(\tilde s_t)_t$ consistent with a distribution with bounded Poincar\'e constant. An interesting open question is if the same bound can be achieved by a computationally efficient algorithm, or if there is a computational-statistical gap.

Finally, we demonstrate through a simple construction that \Cref{prop:it-upper-intro} is essentially tight, i.e. the logarithmic dependence on the Poincar\'e constant cannot be generically improved (if $K < 2^d$):

\begin{proposition}[Informal; see \Cref{prop:it-lower-main}]\label{prop:it-lower-intro}
For any dimension $d$ and Poincar\'e bound $K$, there are distributions $p,q$ on $\RR^d$ with $C_P(p) = C_P(q) \leq K$ and a score function that is $\epsscore^2$-accurate for both, but
\[\TV(p,q) \gtrsim \min\left\{1, \epsscore \sqrt{\min(d, \log K)}\right\}.\]
\end{proposition}

\subsection{Related work}\label{sec:related}

\paragraph{Theory of diffusion models.} Most theoretical analyses of the diffusion sampler incur an error of $\epsscore\sqrt{T}$, via the path-space analysis, where $T$ is the diffusion horizon. The exceptions that we are aware of assume bounds on \emph{exponential moments} of the score error \cite{yang2022convergence,pedrotti2024improved} --- a much stronger assumption than the standard $L^2$ bound, which enables straightforward change-of-measure between the ideal and learned reverse diffusion.\footnote{\cite{pedrotti2024improved} argue that an $L^2$ bound implies an exponential moment bound by clipping the estimated score, but the conversion is lossy and results in a sub-linear dependence on $\epsscore$; their bound also incurs explicit dimension-dependence (even without discretization), and they analyze a modified predictor-corrector algorithm rather than the standard SDE. Moreover, both works \cite{yang2022convergence,pedrotti2024improved} require a log-Sobolev inequality, which is strictly stronger than the Poincar\'e inequality.} At a technical level, these works derive differential inequalities for the divergence $\Dkl{\phat_{T-t}}{p_t}$, an approach that does not seem directly amenable to analyzing specific test statistics.

Several recent works on continuous and discrete diffusion do bound the discrepancy of individual test functions by an adjoint estimate \cite{mimikos2024score,kan2026dimension}, but their focus is not on error accumulation, and indeed they incur horizon-dependence comparable to that of the path-space analysis.

\cite{li2024error} develop a standalone theory of error propagation in diffusion models, but it requires an exact ansatz for the form of the learned score.

\paragraph{Sampling with the estimated vanilla score.} A related question is how well an $L^2$ estimate of the ``vanilla'' score function (algorithmically or information-theoretically) identifies the underlying distribution.\footnote{That is, one is given the score for the data distribution itself, but not for the noised distributions.}  Langevin dynamics with a $\chi^2$-warm start achieves sampling error at most $\epsscore^{1/4} \poly(C_{LS})$, where $C_{LS}$ is the log-Sobolev constant \cite{lee2022convergence}; the error blowup is a consequence of needing to simulate the dynamics for $\poly(C_{LS})$ time. Several recent works provide (algorithmic and information-theoretic) negative results \cite{cao2026robustness,gu2026tractability}.

\paragraph{Sensitivity of generic Markov chains.} There is an extensive literature in probability theory \cite{schweitzer1968perturbation,rudolf2024perturbations} and theoretical computer science \cite{bresler2019stein,gaitonde2025algorithmic,zuckerman2026markov} on how perturbing a Markov chain's transitions affects its $n$-step distributions and its stationary distribution. However, the correct analogies with the diffusion model setting remain unclear. Classical results show that the error accumulation for $n$-step distributions can be mitigated when the Markov chain is fast-mixing, but such results typically require that the perturbations in the transition matrix are small uniformly across the state space \cite{mitrophanov2005sensitivity,rudolf2024perturbations}. Stein's method for bounding the discrepancy between stationary distributions \cite{bresler2019stein} applies even when the perturbations are bounded on average, but it does not directly apply to $n$-step distributions. We believe that there is likely an analogue of our results for generic Markov chains, but we leave its development as an open conceptual problem.

\paragraph{Empirical study of errors in diffusion models.} Error accumulation (also referred to as \emph{exposure bias}) has been observed empirically on datasets such as ImageNet \cite{pmlr-v202-ning23a,li2024error}---though it is unclear whether the primary cause is score error or discretization error---and has motivated investigation of modified training procedures \cite{daras2023consistent}. 

More broadly, it has been widely observed that diffusion models can achieve high perceptual quality while making ``simple'' mistakes---producing physically impossible images \cite{aithal2024understanding} or skewing the likelihood of different image categories compared to the training distributions \cite{perera2023analyzing}. The first type of mistake is generally believed to arise from structural biases in the learned score itself \cite{pmlr-v267-kamb25a,aithal2024understanding,lu2025towards}, but the second type of mistake is less well-understood, and our statistic-specific error bounds provide a potentially explanatory lens. 

\paragraph{Inference-time algorithm design.} Finally, our work can be interpreted as part of a broader agenda on understanding the fundamental limits of designing \emph{inference-time algorithms}, i.e. sampling algorithms with access to powerful yet imperfect oracles motivated by generative modelling---and in particular, understanding how these imperfections accumulate. A recent line of work has investigated this question in the setting of inference-time alignment (see e.g. \cite{pmlr-v267-huang25c,rohatgi2026taming,zhu2026on,jeon2026imperfect}), but the focus is on imperfections in a verifier model rather than the generator.

\subsection{Outline}

In \Cref{sec:setting} we formally introduce the diffusion model setting and the \spectralineq{f}. In \Cref{sec:overview} we overview the proofs of \Cref{thm:observable-intro} and \Cref{cor:poincare-intro}, and in \Cref{sec:overview-it} we overview the proofs of \Cref{prop:it-upper-intro,prop:it-lower-intro}. \Cref{sec:conclusion} concludes with future directions.

We defer formal proofs to the appendices: in \Cref{sec:adjoint} we develop a version of the adjoint method for bounding the discrepancy between two generic SDEs, which we apply in \Cref{sec:ou} to prove \Cref{thm:observable-intro} and \Cref{cor:poincare-intro}. In \Cref{sec:it-lower,sec:projection} we prove \Cref{prop:it-lower-intro} and \Cref{prop:it-upper-intro} respectively. The remaining appendices collect supplementary results and technical lemmas.

\section{Preliminaries}\label{sec:setting}
For a distribution $\mu$ on $\RR^d$, let $L^2(\mu)$ denote the functions $f:\RR^d \to \RR$ with finite squared integral, with inner product $\langle h,k\rangle_{L^2(\mu)}=\int hk\,d\mu$. Similarly let $L^2(\mu;\RR^d)$ denote the square-integrable vector fields $V:\RR^d\to\RR^d$. Define the Sobolev space
\[H^1(\mu) := \{h \in L^2(\mu): \grad h \text{ exists, and lies in } L^2(\mu;\RR^d)\}.\] 
Write $\mathbb E_\mu h=\int h\,d\mu$ and
$\Var_\mu(h)=\mathbb E_\mu(h-\mathbb E_\mu h)^2$. Let  $\|\cdot\|$ denote the standard Euclidean norm. Write $a\vee b := \max\{a,b\}$. Let $I_d$ denote the $d \times d$ identity matrix. Write $f(x) \lesssim g(x)$ to denote $f(x) \leq c g(x)$ for a universal constant $c$. All logarithms are natural.

\subsection{Setting: Ornstein-Uhlenbeck and the learned reverse diffusion}

Throughout the paper, we fix a dimension $d\ge1$, a \emph{horizon} $T > 0$, and a distribution $p$ on $\mathbb R^d$, which we refer to as the \emph{ground truth} law. We follow the theoretical framework in \cite{chen2023sampling}.

The forward diffusion process $(Z_t)_t$ is defined by the Ornstein-Uhlenbeck stochastic differential equation (SDE):
\begin{align}
 dZ_t&=-Z_t\,dt+\sqrt2\,dW_t,\qquad Z_0\sim p
 \label{eq:forward-ou}
\end{align}
where the process $(W_t)_t$ denotes the standard $d$-dimensional Brownian motion. The marginal law of $Z_t$ is
\[p_t = \Law(e^{-t} Z_0 + \sqrt{1 - e^{-2t}}G)\]
where $G\sim\mathcal \normal(0,I_d)$ is independent of $Z_0$. Write $\Zrev_u := Z_{T-u}$. Then $(\Zrev_u)_u$ satisfies the following SDE, which is the time-reversal of the Ornstein-Uhlenbeck SDE:
\[d\Zrev_u = (\Zrev_u + 2\grad \log p_{T-u}(\Zrev_u))du + \sqrt{2} dB_u,\qquad \Zrev_0 \sim p_T\]
where $(B_u)_u$ is standard $d$-dimensional Brownian motion. This SDE provides a schematic for approximate sampling from $p$: first, sample from the standard normal $\gamma := \normal(0,I_d)$ (to which $p_T$ converges as $T \to \infty$); then, follow a discretization of the SDE from $u=0$ to $T$. Both the error due to initializing at $\gamma$ instead of $p_T$, and the error due to discretization, can be controlled by increasing $T$ or decreasing the granularity of discretization \cite{chen2023sampling}. However, this schematic requires exact query access to the score functions $(x,t) \mapsto \grad \log p_t(x)$.

\paragraph{Learned score and learned reverse SDE.} In practice, one has access to a learned score function $(x,t) \mapsto \shat(x,t)$. Throughout the paper, we assume that $\shat$ satisfies the following $L^2$ error bound, which is standard in theoretical analyses of diffusion models \cite{chen2023sampling,chen2023probability,lee2022convergence}, since $\shat$ is typically estimated by minimizing an empirical squared loss.\footnote{Another common theoretical assumption takes a weighted average over times $t$ \cite{chen2023improved,huang2026denoising}, rather than assuming a uniform bound. However, this is likely too weak to obtain any improvement over the path-space analysis, even for ``nice'' distributions.} 

\begin{assumption}\label{assm:score-error}
It holds that for all $0 < t < T$,
\[\ScoreErr{p_t}{\shat_t} := \EE_{p_t} \norm{\shat_t(x) - \grad \log p_t(x)}^2 \leq \epsscore^2.\]
\end{assumption}

We let $(Y_u)_u$ denote the stochastic process defined by the learned reverse SDE, which is obtained by replacing the score function with $\shat$ in the (true) reverse SDE:
\begin{align}
 dY_u&=\bigl(Y_u+2\widehat s_{T-u}(Y_u)\bigr)du+\sqrt2\,dB_u,
 &Y_0&\sim\nu_0,\qquad 0\le u\le T.
 \label{eq:learned-reverse-sde}
\end{align}

We let $\phat_u$ denote the marginal law of $Y_u$, so that the ``output'' of the learned reverse SDE is a sample from $\phat_T$. Formally, any efficient algorithm must use some numerical discretization schema to approximately track the SDE, but we ignore discretization error and focus on bounding the error between $\phat_T$ and $p$; see \Cref{remark:discretization}. We write $\eta_T := \TV(\nu_0, p_T)$ to denote the \emph{initialization error} of the learned SDE.

\begin{remark}[Initialization error]\label{remark:initialization}
With the standard Gaussian initialization $\nu_0 := \normal(0,I_d)$, the standard bound on initialization error \cite{chen2023sampling} is
\[\eta_T \leq \sqrt{\Dkl{p_T}{\normal(0,I_d)}} \leq e^{-T}\sqrt{\Dkl{p}{\normal(0,I_d)}}\]
via Pinsker's inequality and contraction of the Ornstein-Uhlenbeck process; this is the source of the first term in \eqref{eq:standard-girsanov-analysis}, as well as our results (\Cref{thm:observable-intro} and \Cref{cor:poincare-intro}). We do not attempt to tighten the standard dependence on initialization error.
\end{remark}

\subsection{Test statistics and functional inequalities}

A \emph{test statistic} is simply a functional $f:\RR^d \to \RR$ (satisfying the regularity properties described in \Cref{sec:conventions}). As explained in \Cref{sec:contributions}, for a given test statistic $f$, the \spectralineq{f} for distribution $\mu$ relates the transport energy $\TT_\mu(g)$ of perturbation by the centered statistic $g := f-\EE_\mu[f]$, to the variance of the statistic. Classical results from optimal transport theory \cite{benamou2000computational,villani2021topics} imply the following equivalent definitions for $\TT_\mu(g)$:
\begin{equation} \TT_\mu(g) := \inf_{\substack{V \in L^2(\mu;\RR^d): \\ g\mu = -\grad \cdot (\mu V)}}\EE_\mu\norm{V}^2 = \sup_{\substack{h\in H^1(\mu): \\ \EE_\mu\norm{\grad h}^2>0}} \frac{\EE_\mu[gh]^2}{\EE_\mu\norm{\grad h}^2} = \lim_{\delta\to 0} \frac{\WW_2^2((1+\delta g)\mu, \mu)}{\delta^2}.\label{eq:tangent-norm-definitions}\end{equation}
In this paper we use the first and second definitions, not the third, so we formally adopt the first definition and include a proof that the second is equivalent (\Cref{prop:flux-duality}). When a vector field $V$ satisfies $-\grad \cdot (\mu V) = (f-\EE_\mu f)\mu$, we will say that $V$ \emph{realizes perturbation (of $\mu$) by $f$}.

\begin{remark}
The first expression in \eqref{eq:tangent-norm-definitions} is equivalent to the third via the Benamou-Brenier formula \cite{benamou2000computational}: it measures the minimum $L^2$ energy of satisfying the continuity equation for the time-varying density $\delta \mapsto (1+\delta g)\mu$, locally at $\delta=0$. The second expression in \eqref{eq:tangent-norm-definitions} is precisely $\norm{g}_{\dot{H}^{-1}(\mu)}^2$, the squared negative-order weighted homogeneous Sobolev norm of $g$. See e.g. \cite{peyre2018} for further discussion and references.
\end{remark}

\begin{definition}[\spectralineq{f}]\label{def:spectral-ineq}
Let $\mu$ be a distribution on $\RR^d$, and let $f$ be a test statistic with $\Var_\mu(f)>0$. We define
\[C_P(\mu;f) := \frac{\TT_\mu(f-\EE_\mu f)}{\Var_\mu(f)},\]
and we say that $\mu$ satisfies the \spectralineq{f} with constant $C_P(\mu;f)$.
\end{definition}

For completeness, we include the definition of the standard Poincar\'e inequality:

\begin{definition}[Poincar\'e inequality]\label{def:poincare}
Let $\mu$ be a distribution on $\RR^d$. We define 
\[C_P(\mu) := \sup_{f \in H^1(\mu): \EE_\mu\norm{\grad f}^2 > 0} \frac{\Var_\mu(f)}{\EE_\mu\norm{\grad f}^2},\]
and we say that $\mu$ satisfies the Poincar\'e inequality with constant $C_P(\mu)$.
\end{definition}

It is straightforward to show (from the second expression for $\TT_\mu(g)$ in \eqref{eq:tangent-norm-definitions}) that the standard Poincar\'e inequality is equivalent to the \spectralineq{f} holding uniformly over $f$. That is, $C_P(\mu) = \sup_{f:\Var_\mu(f) > 0} C_P(\mu;f)$ (\Cref{lemma:local-global-equivalence}), which justifies our nomenclature. This equivalence is closely related to the \emph{quadratic transportation-variance inequality} \cite{jourdain2012equivalence,ledoux2018remarks,liu2020poincare}.

\subsection{Additional conventions}\label{sec:conventions}

For any distribution $\mu$ on $\RR^d$, we will use $\mu_t$ to denote the pushforward of $\mu$ under the Ornstein-Uhlenbeck process for $t$ time, unless we state otherwise (notably, we deviate from this convention in \Cref{sec:adjoint}, where we consider more general SDEs).

\paragraph{Implicit regularity conditions.} To avoid excess notational overhead, we make the following regularity conditions implicit for the rest of the paper (unless stated otherwise). All functions are assumed to be measurable. Every \emph{distribution} on $\RR^d$ is assumed to have positive, continuously differentiable density. Every \emph{vector field} $V: \RR^d \to \RR^d$ is assumed to be continuously differentiable. Every \emph{test statistic} $f: \RR^d \to \RR$ is assumed to be uniformly continuous, twice continuously differentiable, and square-integrable under the relevant distribution. We assume that the true and approximate score functions $\grad \log p_t(x)$ and $\shat_t(x)$ are Lipschitz in $x$ (uniformly in $t$)\footnote{The condition on $p$ is satisfied if e.g. $p$ is the convolution of a compactly supported distribution with (arbitrarily low-variance) Gaussian noise (\Cref{lemma:lipschitz-score}). We only use it to invoke certain standard stochastic calculus results (\Cref{lemma:feynman-kac,lemma:girsanov}) and it can likely be relaxed.} and continuous in $(x,t)$. None of our results have quantitative dependence on the constants implicit in these conditions.

\section{Technical overview I: stability of the learned SDE}\label{sec:overview}

In this section, we explain the key elements in the proof of \Cref{thm:observable-intro} (as well as \Cref{cor:poincare-intro}), expanding on the intuition given in \Cref{sec:contributions}. For simplicity, we assume for now that there is no initialization error (i.e. $\nu_0 = p_T$), to focus on the key techniques.

Fix a bounded test statistic $f$, i.e. a measurable function $f: \RR^d \to [0,1]$. Let $(f_t)_t$ denote the posterior statistics $f_t(z) := \EE[f(Z_0) \mid{} Z_t = z]$ under the true stochastic process. 

\paragraph{The adjoint method (\Cref{sec:adjoint}).} The starting point for the proof of \Cref{thm:observable-intro} is the following bound on the discrepancy under $f$ in terms of the Dirichlet energies of the posterior statistics (\Cref{lem:whole-comparison}):
\begin{equation}
 |\mathbb E_{\phat_T}f-\mathbb E_p f|
 \leq 2\epsscore\int_0^T
\sqrt{\EE_{p_t}\norm{\grad f_t}^2}\,\dd t + \epsscore^2 T.
\label{eq:adjoint-overview}
\end{equation}
This bound is not specific to the diffusion process; it applies to the terminal marginals of any two SDEs with the same initial condition and a bound of $4\epsscore^2$ on the $L^2$ difference in their drifts. It is a variant of the \emph{adjoint method} for bounding the accumulation of errors due to discretization \cite{talay1990expansion} or learning \cite{mimikos2024score,kan2026dimension};\footnote{To explain the nomenclature, $u(x,t) = f_t(x)$ solves the \emph{adjoint equation} for the stochastic process.} the key distinction from prior applications \cite{mimikos2024score,kan2026dimension} is that the bound in \eqref{eq:adjoint-overview} \emph{does not depend on the learned process},\footnote{In contrast, the more straightforward estimate \cite{mimikos2024score,kan2026dimension} gives the bound $\epsscore\int_0^T
\sqrt{\EE_{\phat_t}\norm{\grad f_t}^2}\,\dd t$, or symmetrically $\epsscore\int_0^T
\sqrt{\EE_{p_t}\norm{\grad \wh{f}_t}^2}\,\dd t$, where $\wh f_t$ is the posterior statistic under the learned stochastic process.} so it can (in principle) be directly controlled in terms of $f$ and $p$.

It is illustrative to see how one can recover the standard bound $ |\mathbb E_{\widehat p_T}f-\mathbb E_p f|
 \lesssim \epsscore \sqrt{T}$ using \eqref{eq:adjoint-overview}, before discussing how to beat it. Indeed, the Dirichlet energy quantifies the \emph{dissipation of variance} of the posterior statistic (\Cref{lem:posterior-variance}):
 \begin{equation} \frac{\dd}{\dd t} \Var_{p_t}(f_t) = -2 \EE_{p_t} \norm{\grad f_t}^2.\label{eq:dissipation-overview}\end{equation}
Thus, $\Var_{p_t}(f_t)$ serves as a potential function for the Dirichlet energy. Since $f_t \in [0,1]$ almost surely, this potential is always at most $1$, so $\int_0^T \EE_{p_t} \norm{\grad f}^2 \, \dd t \leq 1$. By Cauchy-Schwartz, one gets
\[\int_0^T \sqrt{\EE_{p_t}\norm{\grad f}^2} \, \dd t \leq \sqrt{T},\]
which, together with \eqref{eq:adjoint-overview}, implies that $ |\mathbb E_{\widehat p_T}f-\mathbb E_p f|
 \lesssim \epsscore \sqrt{T}$. Since this holds for any bounded statistic $f$, it follows that $\TV(\phat_T, p) \lesssim \epsscore\sqrt{T}$ as well---which is the same bound obtained by the standard path-space analysis \eqref{eq:standard-girsanov-analysis}.

 However, the above proof via \eqref{eq:adjoint-overview} provides additional insight into when the $\sqrt{T}$-dependence can be improved. Cauchy-Schwartz is (asymptotically) tight when $\EE_{p_t} \norm{\grad f_t}^2$ is roughly constant (proportional to $1/T$) across $0 \leq t \leq T$. In this situation, by \eqref{eq:dissipation-overview}, $\Var_{p_t}(f_t)$ decays roughly linearly from $t=0$ to $T$. To beat Cauchy-Schwarz, it suffices to show that the variance decays \emph{super-linearly}. 
 

 \subsection{Improved bounds under a global Poincar\'e inequality (\Cref{cor:poincare-intro})} 
\label{sec:poincare-overview} 
 Up to now, the analysis has applied to generic SDEs; establishing variance decay for the posterior statistics will require exploiting the structure of the Ornstein-Uhlenbeck process. As outlined in \Cref{sec:contributions}, combining \eqref{eq:dissipation-overview} with the Poincar\'e inequality gives the following differential inequality for the variance:
 \begin{equation}\frac{\dd}{\dd t} \Var_{p_t}(f_t) = -2\EE_{p_t}\norm{\grad f_t}^2 \leq -\frac{2}{C_P(p_t)} \Var_{p_t}(f_t).\label{eq:differential-inequality-overview}\end{equation}
 By the smoothing property of the Ornstein-Uhlenbeck process, we have $C_P(p_t) \leq e^{-2t}C_P(p)+1-e^{-2t}$ for all $t\geq 0$. Substituting into \eqref{eq:differential-inequality-overview} and solving the differential inequality shows that the variance decays exponentially:
 \[\Var_{p_t}(f_t) \lesssim e^{-2t}\max(1,C_P(p))\Var_p(f).\]
As sketched above, this must suffice to beat Cauchy-Schwarz; quantitatively, we can bound the integral in \eqref{eq:adjoint-overview} by applying Cauchy-Schwarz within short intervals:
\[\int_0^T \sqrt{\EE_{p_t}\norm{\grad f_t}^2}\,\dd t \leq \sum_{n=0}^{\lfloor T\rfloor} \sqrt{\int_{n}^{n+1} \EE_{p_t} \norm{\grad f_t}^2 \,\dd t} \leq \sum_{n=0}^{\lfloor T\rfloor} \sqrt{\Var_{p_n}(f_n)} \lesssim \sqrt{1+C_P(p)}.\]
A slightly more careful calculation---treating the interval $[0,\log(1+C_P(p))]$ separately---strengthens to bound to $\sqrt{\log(1+C_P(p))}$. By substituting into \eqref{eq:adjoint-overview} and supremizing over all bounded $f$, this suffices to prove \Cref{cor:poincare-intro}, though of course we formally obtain it as a corollary of \Cref{thm:observable-intro} (discussed next).

\subsection{Beyond global Poincar\'e: analyzing a fixed statistic (\Cref{thm:observable-intro})} 

We now explain why it is possible to circumvent the global Poincar\'e assumption when, as in \Cref{thm:observable-intro}, we only care about a specific test statistic $f$. The key point is that the above argument is already specified to a single $f$: to bound the discrepancy under $f$, it only invokes the Poincar\'e inequality for $p_t$ with the functional $f_t$. But if we only make an assumption on the functional $f$ (namely, the \spectralineq{f}), how do we convert this into the necessary bounds on all posterior statistics $f_t$?

\paragraph{From a transport map to a Poincar\'e-type bound.} First, we explain why the \spectralineq{f} implies the necessary bound on $f$ itself. Intuitively, this is clear from \Cref{def:spectral-ineq} and \eqref{eq:tangent-norm-definitions}: when $C_P(p;f)$ is bounded, there exists a low-energy velocity field $V$ that drives $p$ to $(1+\delta g)p$ (where $g := f-\EE_p f$), meaning that the variation in $f$ must be within well-connected components of $p$, which should imply that $\Var_p(f)$ can be bounded in terms of $\EE_p\norm{\grad f}^2$.

This intuition can easily be made precise using the second, dual definition of the transport energy $\TT_p(g)$ (see \Cref{prop:flux-duality} for the proof of equivalence): for every functional $h$, it holds that
\[\EE_p[gh]^2 \leq \TT_p(g) \EE_p \norm{\grad h}^2.\]
Taking $h := g = f-\EE_p f$ and using the \spectralineq{f} gives
\[\Var_p(f)^2 \leq C_P(p;f)\Var_p(f)\EE_p\norm{\grad f}^2,\]
which simplifies to the standard Poincar\'e-type bound $\Var_p(f) \leq C_P(p;f)\EE_p\norm{\grad f}^2$.

\paragraph{Propagating to posterior statistics.} Next, we argue that, under the Ornstein-Uhlenbeck diffusion, the perturbative velocity field itself propagates to the posterior statistics. In particular, if vector field $V$ realizes perturbation of $p$ by $f$, then
\[V_t(y) := e^{-t}\EE[V(Z_0) \mid{} Z_t = y]\]
realizes perturbation of $p_t$ by $f_t$ (\Cref{lem:ou-flux}). Since $\EE_{p_t}\norm{V_t}^2 \leq e^{-2t}\EE_p\norm{V}^2$, we get that
\[\TT_{p_t}(f_t-\EE_{p_t} f_t) \leq e^{-2t} \TT_p(f-\EE_p f).\]
The same argument as above then gives
\[\Var_{p_t}(f_t)^2 \leq e^{-2t} C_P(p;f)\Var_p(f) \EE_{p_t}\norm{\grad f_t}^2,\]
and thus (since $\Var_p(f) \leq 1$),
\[\Var_{p_t}(f_t) \leq e^{-t}\sqrt{C_P(p;f)\EE_{p_t}\norm{\grad f_t}^2}.\]
This is not quite the standard Poincar\'e-type bound, but it's sufficient to implement the argument outlined in \Cref{sec:poincare-overview}, this time without the global Poincar\'e assumption. The differential inequality for $\Var_{p_t}(f_t)$ is weaker than \eqref{eq:differential-inequality-overview} but still implies exponential decay (\Cref{lemma:variance-decay}); using this decay, the integral can be bounded by $O(\sqrt{\log(1+C_P(p;f))})$ (\Cref{lemma:dirichlet-integral-bound}) as needed for \Cref{thm:observable-intro}. See \Cref{sec:ou} for the full proof.

\section{Technical overview II: information-theoretic limits}\label{sec:overview-it}
In this section, we provide an overview of the proofs of \Cref{prop:it-lower-intro,prop:it-upper-intro}, which together precisely establish the information-theoretic limits of sampling with approximate score functions under the Poincar\'e inequality. We start with the information-theoretic lower bound, \Cref{prop:it-lower-intro}, as the proof idea contains useful intuition for analyzing the optimal algorithm.

\subsection{Information-theoretic lower bound (\Cref{prop:it-lower-intro})}\label{sec:it-lower-overview}

Suppose we know the score functions for a distribution $p$ up to an $L^2$ error of $\epsscore^2$. How precisely can we identify $p$? 

A natural place to start examining this question is with univariate Gaussian distributions, for which score errors can be computed explicitly, and total variation distance can be asymptotically bounded. Suppose that $p := \normal(0,\sigma^2)$ and $q := \normal(\mu,\sigma^2)$ for some $\sigma \geq 2$ and $\mu>0$; we will think of $\grad \log q_t$ as the ``estimated score'' for $p_t$. Then for any $t \geq 0$,
\[\sqrt{\ScoreErr{p_t}{\grad \log q_t}} = \frac{e^{-t}\mu}{e^{-2t}\sigma^2 + 1 - e^{-2t}} = \frac{\mu}{e^{-t}\sigma^2 + e^t - e^{-t}}.\]
The minimum value of the denominator (across choices of $t$) is $2\sqrt{\sigma^2-1} = \Theta(\sigma)$, and hence
\[\sup_{t \geq 0} \sqrt{\ScoreErr{p_t}{\grad \log q_t}} = \Theta(\mu/\sigma).\]
Thus, the score error is bounded by $\epsscore^2$ uniformly in $t$, so long as $\mu \leq O(\epsscore \sigma)$. In this regime, standard bounds give $\TV(p,q) \leq O(\epsscore)$. Hence, there is no ``accumulation of error.'' It is natural to wonder whether this argument can be made generic: perhaps knowing the annealed scores up to error $\epsscore^2$ determines \emph{any} distribution up to total variation distance $O(\epsscore)$.

Unfortunately, this is false, and while univariate Gaussian distributions do not provide a counterexample, they do already provide an avenue towards one. In the above example, while there exists a time $\tcrit = \tstar_{\mu,\sigma}$ at which the score error has magnitude $\Theta(\mu^2/\sigma^2)$, the error decays rapidly for both $t \gg \tcrit$ and $t \ll \tcrit$. In other words, the mis-identification of the mean of $p$ contributes to the score error almost exclusively in a small window around this $\tcrit$. Intuitively, this is the window where the noised distributions $p_t$ and $q_t$ have started to overlap, but do not yet coincide. Increasing $\sigma$ shifts the window to larger $t$.

\paragraph{Multi-scale distributions.} These observations suggest constructing distributions $p,q$ where different discrepancies between the distributions are visible in the annealed score at different windows of time. Each discrepancy will contribute to the distance between $p$ and $q$, but the total score error at any fixed time $t$ will never be too large.

There are several ways one can imagine such multi-scale distributions. In the construction for \Cref{prop:it-lower-intro}, we take $p$ and $q$ to be multivariate Gaussian distributions where each dimension is in a different scale. To enforce the Poincar\'e bound, we need the largest scale (i.e. the largest eigenvalue of the shared covariance matrix) to be at most $K$. To ensure that each dimension contributes to the score error in a different window of time, we need the eigenvalues to be an exponential sequence bounded below by $1$. Thus, we can afford $O(\log K)$ different scales, which yields the desired $O(\sqrt{\log K})$ blowup in $\TV(p,q)$. See \Cref{sec:it-lower} for the full proof.

\subsection{Information-theoretically optimal but inefficient algorithm (\Cref{prop:it-upper-intro})}

Ignoring computational issues, the optimal sampler $\cA$ is easy to describe: given the approximate score function $\shat$, $\cA(\shat)$ picks any distribution $q$ such that (1) $q$ has bounded Poincar\'e constant, and (2) $\shat$ has low score error relative to $q$. Without further information about the true distribution $p$, this estimator clearly achieves optimal (i.e., minimax) total variation error up to constant factors. Thus, proving \Cref{prop:it-upper-main} reduces to showing that, for any distributions $p,q$ with $\max\{C_P(p), C_P(q)\} \leq K$ and 
\[\sup_{t \geq 0} \max\{\ScoreErr{p_t}{\shat_t}, \ScoreErr{q_t}{\shat_t}\} \leq \epsscore^2,\]
the total variation distance between $p$ and $q$ is at most $O(\epsscore \sqrt{\log(K)})$.

\paragraph{Total variation bounds via the reverse SDE.} Although the information-theoretic method $\cA$ described above cannot (to our knowledge) be implemented via reverse diffusion, an idealized version of the reverse diffusion SDE nevertheless provides an approach for analyzing the method. Specifically, suppose we knew that $p_{\tstar} = q_{\tstar}$ for some time $\tstar>0$. Consider the approximate reverse diffusion process initialized with $p_{\tstar} = q_{\tstar}$ and using score estimate $\shat$; let the marginal law of the output be $\wh\mu$. Girsanov's theorem implies that $\TV(\wh\mu, p) \leq \epsscore\sqrt{\tstar}$ and $\TV(\wh\mu, q) \leq \epsscore\sqrt{\tstar}$, so by the triangle inequality, $\TV(p,q) \leq 2\epsscore\sqrt{\tstar}$. More generally, even if $p_{\tstar} \neq q_{\tstar}$, the same idea shows that (\Cref{lem:final-girsanov})
\[\TV(p,q) \leq 2\epsscore\sqrt{\tstar} + \TV(p_{\tstar}, q_{\tstar}).\]
Thus, it suffices to show that $\TV(p_{\tstar}, q_{\tstar}) \leq O(\epsscore)$ for some $\tstar \leq O(\log K)$.

\paragraph{Multimodality, diffusion, and the critical time.} We define the initialization time to be $\tstar := \frac{1}{2} \log(K)$. Since the Ornstein-Uhlenbeck process exponentially attenuates Poincar\'e constants, we know that $\max\{C_P(p_{\tstar}), C_P(q_{\tstar})\} \leq O(1)$. In other words, $p_{\tstar}$ and $q_{\tstar}$ can be thought of as roughly unimodal.

Suppose, for intuition, that $\shat_{\tstar}$ is exactly the score of some other distribution $r$. It is tempting to try to directly control $\TV(p_{\tstar},r)$ and $\TV(q_{\tstar},r)$ in terms of the respective score errors. Unfortunately, such bounds require an isoperimetric assumption on $r$ \cite{koehler2022statistical}. Indeed, if $r$ is a mixture of $p_{\tstar}$ and some other well-separated component, then $\ScoreErr{p_{\tstar}}{\grad \log r}$ can be arbitrarily small even though $\TV(p_{\tstar},r) = \Omega(1)$. More specifically, there is the following obstacle: if $p_{\tstar}$ and $q_{\tstar}$ are well-separated, and $r = \frac{p_{\tstar}+q_{\tstar}}{2}$, then we can simultaneously have 
\[\max\{\ScoreErr{p_{\tstar}}{\grad \log r},\ScoreErr{q_{\tstar}}{\grad \log r}\} \leq \epsscore^2\]
and yet $\TV(p_{\tstar},q_{\tstar}) = \Omega(1)$, even if $p_{\tstar}$ and $q_{\tstar}$ both have bounded Poincar\'e constant.

However, this is \emph{not} a counterexample to the claim that $\TV(p_{\tstar},q_{\tstar}) \leq O(\epsscore)$. It merely shows that to prove the claim, we must leverage the score error bounds at \emph{later} times $t > \tstar$, not just the bounds at time $\tstar$. Moreover, since the above obstacle appears when $p_{\tstar}$ and $q_{\tstar}$ are well-separated, this (along with the intuition from the Gaussian case in \Cref{sec:it-lower-overview}) motivates considering the \emph{first time $t$ at which $p_t$ and $q_t$ have non-trivial overlap}. We define the \emph{critical time} $\tcrit$ by
\[\tcrit := \inf\{t \geq \tstar: \TV(p_t, q_t) \leq 8\epsscore\}.\]
We bound $\TV(p_{\tstar},q_{\tstar})$ by showing that it's \emph{impossible} to have $\tcrit > \tstar$.\footnote{The critical time is finite since, under mild regularity, the Ornstein-Uhlenbeck process ensures that $\TV(p_t, q_t) \to 0$ (\Cref{lemma:tv-continuity-convergence})} Intuitively, since $p_{\tcrit}$ and $q_{\tcrit}$ have substantial overlap, the above obstacle is ruled out. Formally, we show that it suffices to consider the case where $\shat_{\tcrit}$ is the score of the mixture $r := \frac{p_{\tcrit} + q_{\tcrit}}{2}$ (\Cref{lem:mixture-score}). Since $p_{\tcrit}$ and $q_{\tcrit}$ have substantial overlap and individually have good Poincar\'e constants, we can bound the Poincar\'e constant of $r$ (\Cref{lem:mixture-poincare}). Thus, $\TV(p_{\tcrit},r)$ and $\TV(q_{\tcrit},r)$ can be bounded in terms of their score errors (\Cref{lem:fisher-chi}), and we get
\[\TV(p_{\tcrit},q_{\tcrit}) \leq \TV(p_{\tcrit},r) + \TV(q_{\tcrit},r) \leq 7\epsscore,\]
which contradicts the choice of $\tcrit$ as the \emph{first} time at which $\TV(p_t,q_t) \leq 8\epsscore$. This shows that necessarily $\tcrit = \tstar$, which, together with the Girsanov bound, completes the proof of \Cref{prop:it-upper-intro}. See \Cref{sec:projection} for the full proof.

\section{Conclusion} \label{sec:conclusion}

In this work, we provide a finer-grained study of how errors in learned scores propagate through diffusion sampling. In particular, we study the sensitivity of the continuous reverse SDE to score errors, independent of numerical discretization, and we prove (algorithm-independent) information-theoretic limits.

By providing guarantees on how well we can match individual test statistics, we identify geometric
conditions under which score errors accumulate more benignly than standard path-space bounds suggest. In particular, a Poincar\'e inequality suffices to get guarantees in total variation distance that are independent of the dimension. Moreover, our broader framework shows that selected statistics can enjoy similar guarantees even without such global geometric control.

Our results suggest several directions for further work. For example, which other structural properties of the data distribution, and which other error metrics permit benign error propagation? A concrete algorithmic question concerns the gap between the information-theoretic lower bound on error accumulation, and what we can do efficiently. In particular, for sufficiently small $\epsscore$, when the higher-order remainder is negligible, our sampling guarantees match the information-theoretic lower bounds.
Can efficient algorithms attain the information-theoretic rate throughout the remaining range of $\epsscore$, or is there a computational complexity or query complexity barrier to doing so?

\paragraph{AI methodology.} We used AI models to assist with proof development for the main results through repeated interactions and author feedback. The information-theoretic results were developed earlier in the project, without the use of AI. No AI was used in specifying the overall research direction, though it was helpful in refining the definition of the \spectralineq{f}, and identifying its connections with the existing literature.\footnote{Specifically, we originally sought an error bound for balanced test statistics on mixture distributions (\Cref{cor:balanced-main}). We then asked an AI model to identify a ``spectral'' condition that makes sense for arbitrary distributions. It came up with one version of \Cref{def:spectral-ineq}, and we elucidated interpretations of this condition via further interactions with AI models.}
All proofs have been verified by the authors.

\paragraph{Acknowledgments.} AM is supported in part by a Microsoft Trustworthy AI Grant, NSF award CCF-2430381, ONR grant N00014-22-1-2339, and a David and Lucile Packard Fellowship. AR is supported in part by NSF awards IIS-2211907, CCF-2238523, IIS-2403275, an Amazon Research
Award, ONR award N000142512124, a Google Research Scholar Award, and an OpenAI Superalignment Fast Grant. DR is supported by a Miller Postdoctoral Fellowship. GG is supported by the Onassis Foundation and the A.G. Leventis Foundation.

\bibliographystyle{alpha}
\bibliography{bib}

\clearpage
\appendix
\crefalias{section}{appendix}
\crefname{appendix}{appendix}{appendices}
\Crefname{appendix}{Appendix}{Appendices}
\titleformat{\section}{\large\bfseries}{Appendix \thesection}{0.6em}{}

\section{Stability of generic SDEs via the adjoint method}\label{sec:adjoint}
In this section, we consider a generalized setting with two arbitrary SDEs, and we introduce a version of the \emph{adjoint method} for bounding the accumulated error between their marginals. In later sections, we will specify these results to analyze the error in the learned reverse diffusion SDE.

\paragraph{Setting for this section.} Formally, throughout this section we let $(X_t)_{t\in[0,T]}$, $(\wh X_t)_{t\in[0,T]}$ be stochastic processes on $\RR^d$, defined by the following SDEs:
\[dX_t = b(X_t, t) dt + \sqrt{2} dW_t,\]
\[d\wh X_t = \wh b(\wh X_t, t) dt + \sqrt{2} dW_t,\]
with initial condition $X_0 = \wh X_0 \sim q_0$. Here, $(W_t)_t$ is standard $d$-dimensional Brownian motion, and $b$, $\wh b: \RR^d \times [0, T]\to\RR^d$ are arbitrary jointly continuous drift functions that are Lipschitz in space, uniformly in time. We let $(q_t)_t$, $(\wh q_t)_t$ denote the respective marginal laws (so that $\wh q_0 = q_0$).

We also fix some bounded test statistic $f: \RR^d \to [0,1]$, and let $f_t$ denote the following posterior (also known as Feynman-Kac solution), defined with respect to the ``true'' process $(X_t)_t$:
\[f_t(x) := \EE[f(X_T) \mid{} X_t = x].\]

\begin{lemma}\label{lem:whole-comparison}
Suppose that
\begin{equation}\sup_{0 < t < T} \EE_{x\sim q_t} \norm{b(x,t) - \wh b(x,t)}^2 \leq \epsilon^2.\label{eq:drift-error}\end{equation}
Then 
\[|\EE_{q_T}[f] - \EE_{\wh q_T}[f]| \leq \epsilon \int_0^T \sqrt{\EE_{q_t} \norm{\grad f_t}^2} \, \dd t + \frac{\epsilon^2}{4} T.\]
\end{lemma}

\begin{proof}
Let $Q$, $\wh Q$ denote the respective path laws of $X=(X_t)_{t=0}^T$ and $\wh X = (\wh X_t)_{t=0}^T$. By \Cref{lemma:log-change-of-measure} and the fact that $0 \leq f \leq 1$, we have
\begin{equation} \EE[f(\wh X_t)] - \EE[f(X_T)] \leq \EE\left[f(X_T)\log(\wh Q(X)/Q(X))\right] + \Dkl{Q}{\wh Q}.\label{eq:magic-log-bound}\end{equation}
By the Lipschitz regularity of the drifts $b, \wh b$, we may apply Girsanov's theorem (\Cref{lemma:girsanov}), which gives that the following equality holds $Q$-almost-surely:
\begin{equation} \log \frac{\wh Q(X)}{Q(X)} = \frac{1}{\sqrt{2}}\int_0^{T} \langle \wh b(X_t, t) - b(X_t, t), \dd W_t\rangle - \frac{1}{4}\int_0^{T} \norm{\wh b(X_t, t) - b(X_t, t)}^2 \dd t.\label{eq:girsanov}\end{equation}
Taking expectation under $Q$ gives $\Dkl{Q}{\wh Q} \leq \frac{1}{4}\epsilon^2 T$. It remains to bound the first term on the RHS of \eqref{eq:magic-log-bound}, which we do by writing each of $f(X_T)$ and $\log(\wh Q(X)/Q(X))$ as an It\^o integral and then applying It\^o's isometry. 

First, by \eqref{eq:girsanov}, dropping the non-positive term, we have
\begin{equation}\log\frac{\wh Q(X)}{Q(X)} \leq \frac{1}{\sqrt{2}}\int_0^T \langle \wh b(X_t,t) - b(X_t,t), dW_t\rangle.\label{eq:girsanov-ub}\end{equation}
Second, by the Lipschitz regularity of $b$ and uniform continuity of $f$ (see \Cref{sec:conventions}), the Feynman-Kac equivalence applies to $(f_t)_t$ (\Cref{lemma:feynman-kac}). That is, it solves the following partial differential equation, known as the \emph{backward Kolmogorov equation} for $(X_t)_t$:
\[\partial_t f_t(x) + \langle b(x,t), \grad f_t(x)\rangle + \laplace f_t(x) = 0, \qquad \forall (x,t) \in \RR^d \times (0,T).\]
It\^o's formula (\Cref{lemma:ito-formula}) applied to the time-varying function $g(x,t) = f_t(x)$ therefore simplifies as follows:
\begin{align}
f_T(X_t) - f_0(X_0)
&= \int_0^T (\partial_t f_t(X_t) + \langle b(X_t,t),\grad f_t(X_t)\rangle + \laplace f_t(X_t))\,\dd t + \sqrt{2}\int_0^T \langle \grad f_t(X_t), dW_t\rangle \nonumber \\ 
&= \sqrt{2}\int_0^T \langle \grad f_t(X_t), dW_t\rangle.\label{eq:ito-ft}
\end{align}
Substituting \eqref{eq:girsanov-ub} and \eqref{eq:ito-ft} into the first term on the RHS of \eqref{eq:magic-log-bound}, we get
\begin{align*}
\EE[f(X_T) \log(\wh Q(X)/Q(X))]
&\leq \frac{1}{\sqrt{2}}\EE\left[f(X_T) \int_0^T \langle \wh b(X_t,t) - b(X_t,t), dW_t\rangle\right] \\ 
&= \frac{1}{\sqrt{2}}\EE\left[\left(f_0(X_0) + \sqrt{2}\int_0^T \langle \grad f_t(X_t), dW_t\rangle\right) \int_0^T \langle \wh b(X_t,t) - b(X_t,t), dW_t\rangle\right] \\ 
&= \EE\left[\int_0^T \langle \grad f_t(X_t), \wh b(X_t, t) - b(X_t,t)\rangle \, \dd t\right]
\end{align*}
where the final equality is by the inner product form of It\^o's isometry (\Cref{lemma:ito-isometry}). Applying Cauchy-Schwarz and \eqref{eq:drift-error} gives
\[\EE[f(X_T) \log(\wh Q(X)/Q(X))] \leq \epsilon\int_0^T \sqrt{\EE_{q_t}[\norm{\grad f_t}^2]}\,\dd t.\]
Substituting into \eqref{eq:magic-log-bound} produces one side of the claimed bound:
\[\EE_{\wh q_T}[f] - \EE_{q_T}[f] \leq \epsilon\int_0^T \sqrt{\EE_{q_t}[\norm{\grad f_t}^2]}\,\dd t + \frac{\epsilon^2}{4} T.\]
Applying the same argument to the statistic $1-f$ completes the proof.
\end{proof}

\begin{lemma}[Posterior variance identity]\label{lem:posterior-variance}
In the setting of the previous lemma, it holds for all $0 < t < T$ that
\begin{align}
 \frac{d}{dt}\Var_{q_t}(f_t)
 &=2\EE_{q_t}\|\nabla f_t\|^2.
 \label{eq:variance-differential-identity}
\end{align}
\end{lemma}
\begin{proof}
As in the proof of the previous lemma, It\^o's formula gives for any $t$ that
\[f_t(X_t) - f_0(X_0) = \sqrt{2} \int_0^t \langle \grad f_s(X_s), dW_s\rangle.\]
Since $\EE[f_t(X_t)] = \EE[f_0(X_0)]$, we can rearrange to get
\[f_t(X_t) - \EE[f_t(X_t)] = f_0(X_0) - \EE[f_0(X_0)] + \sqrt{2} \int_0^t \langle \grad f_s(X_s), dW_s\rangle.\]
Thus, by It\^o's isometry,
\[\EE[(f_t(X_t) - \EE[f_t(X_t)])^2] = \EE[(f_0(X_0) - \EE[f_0(X_0)])^2] + 2\int_0^t \EE[\norm{\grad f_s}^2] \, \dd s.\]
Differentiating yields the claimed result.
\end{proof}

\paragraph{Implications of the adjoint method.} Since any bounded test statistic has bounded variance, \Cref{lem:posterior-variance} directly implies that
\[\int_0^T \EE_{q_t} \norm{\grad f_t}^2 \, \dd t \leq 1.\]
One could apply Cauchy-Schwarz to this bound, and then invoke \Cref{lem:whole-comparison}, to get
\[|\EE_{\wh q_T}f - \EE_{q_T} f| \leq \epsilon\sqrt{T} + \frac{\epsilon^2}{4}T.\]
However, this is no better than the direct approach of applying Girsanov's theorem and Pinsker's inequality to the path laws. The benefit of the adjoint method only comes when we are able to show that the Dirichlet energies $\EE_{q_t} \norm{\grad f_t}^2$ decay as $t$ increases (or, more generally, are non-uniform in time).

\section{Stability of the reverse diffusion process}\label{sec:ou}
In this section, we apply the generic results developed in the previous section to the specific setting of the reverse diffusion process. Specifically, the main goal of this section is to prove \Cref{thm:observable-intro} (formally restated below as \Cref{thm:observable-main}), which bounds the discrepancy in a single test statistic $f$ under the \spectralineq{f}. We also prove \Cref{cor:poincare-intro} (restated as \Cref{cor:poincare-main}), which bounds total variation distance under the Poincar\'e inequality, and \Cref{cor:balanced-main}, which applies \Cref{thm:observable-intro} to mixture distributions.

\paragraph{Notation.} We recall the notation from \Cref{sec:setting}: $(Z_t)_t$ denotes the true Ornstein-Uhlenbeck diffusion process, with marginals $p_t$, and $(Y_t)_t$ denotes the learned reverse process, with marginals $\phat_t$. Note that $\phat_T$ is the output law of the learned process, and it is an approximation to $p = p_0$ (not $p_T$). We let $\eta_T := \TV(\phat_0, p_T)$ denote the initialization error. For any statistic $f: \RR^d \to [0,1]$ and $t \in [0,T]$, we let $f_t$ denote the posterior statistic
\[f_t(y) := \EE[f(Z_0) \mid{} Z_t = y].\]
Note the reversal of time (relative to the notation in \Cref{sec:adjoint}). In this notation, \Cref{lem:posterior-variance} gives
\begin{equation}\frac{\dd}{\dd t} \Var_{p_t}(f_t) = -2\EE_{p_t} \norm{\grad f_t}^2.\label{eq:dissipation}\end{equation}
That is, the variance of $f_t$ decays as $t$ increases, at a rate governed by the Dirichlet energy of $f_t$.

\begin{theorem}\label{thm:observable-main}
Let $f:\mathbb R^d\to[0,1]$ be a bounded test statistic for $p$ with positive variance. Write $C_f := \max(C_P(p;f), 1)$. Then the sampling law $\phat_T$, induced by the learned reverse SDE, satisfies
\begin{align}
 |\mathbb E_{\widehat p_T}f-\mathbb E_p f|
 &\le\eta_T+2\epsscore\sqrt{1+\log(C_f)}
       +\epsscore^2T.
 \label{eq:main-whole-observable-bound}
\end{align}
\end{theorem}

\Cref{thm:observable-intro} follows immediately from \Cref{thm:observable-main} by taking the standard Gaussian initialization $\nu_0 := \normal(0,I_d)$ and bounding the initialization error $\eta_T$ accordingly; see \Cref{remark:initialization}.

\paragraph{Outline of section.} To prove \Cref{thm:observable-main}, we must bound $|\EE_p[f] - \EE_{\phat_T}[f]|$, using the \spectralineq{f}. By the adjoint method (\Cref{lem:whole-comparison}), it suffices to bound the Dirichlet energy integrated over the posterior statistics $f_t$:
\begin{equation} \int_0^T \sqrt{\EE_{p_t} \norm{\grad f_t}^2}\, \dd t.\label{eq:dirichlet-integral-section-outline}\end{equation}
As outlined in \Cref{sec:overview}, we do this by first showing that the \spectralineq{f} propagates to a ``square-root Poincar\'e''-type bound on the posterior statistics (\Cref{sec:propagation}). We then use this bound together with \eqref{eq:dissipation} to show that $\Var_{p_t}(f_t)$ must decay exponentially in $t$ (as opposed to linearly, which would make \eqref{eq:dirichlet-integral-section-outline} scale with $\sqrt{T}$). This enables a horizon-independent bound on the integral (\Cref{sec:variance-decay}). We can then easily conclude the proofs of \Cref{thm:observable-main}, \Cref{cor:poincare-main}, and \Cref{cor:balanced-main} (\Cref{sec:main-proofs}).

\subsection{Propagation to posterior statistics}\label{sec:propagation}

The following lemma shows that the \spectralineq{f} propagates to a ``square-root Poincar\'e''-type bound for the posterior statistics $f_t$:

\begin{lemma}[Square-root Poincar\'e for posterior statistics]\label{lemma:sqrt-poincare}
Let $f:\RR^d \to \RR$ be a test statistic. Then, for every $0 < t < T$, it holds that
\[\Var_{p_t}(f_t) \leq e^{-t} \sqrt{C_P(p;f) \Var_p(f) \EE_{p_t}\norm{\grad f_t}^2}.\]
\end{lemma}

\begin{proof}
Without loss of generality assume that $\EE_p f = 0$, so that $\EE_{p_t} f_t = 0$ for all $t$. Let $V$ be any vector field that realizes perturbation of $p$ by $f$. By \Cref{lem:ou-flux}, there is a field $V_t$ with $\EE_{p_t} \norm{V_t}^2 \leq e^{-2t} \EE_p \norm{V}^2$ that realizes perturbation of $p_t$ by $f_t$. By \Cref{lemma:realizable-field-equivalence}, we have
\[\EE_{p_t}[f_t^2] \leq \sqrt{\EE_{p_t} \norm{V_t}^2 \EE_{p_t} \norm{\grad f_t}^2} \leq e^{-t}\sqrt{\EE_{p} \norm{V}^2 \EE_{p_t} \norm{\grad f_t}^2}.\]
Taking the infimum over all such $V$ and applying the definition of $C_P(p;f)$ (\Cref{def:spectral-ineq}) completes the proof.
\end{proof}

Key to the above proof was the following lemma showing that a vector field realizing a perturbation by $f$ actually propagates to the posterior statistics.

\begin{lemma}[Propagation of realizability]\label{lem:ou-flux}
Let $f: \RR^d \to \RR$ be a test statistic with $\EE_p f = 0$. If $V$ realizes perturbation of $p$ by $f$, then
\begin{align}
 V_t(y)=e^{-t}\mathbb E[V(Z_0)\mid Z_t=y]
 \label{eq:propagated-flux}
\end{align}
realizes perturbation of $p_t$ by $f_t$. Moreover, $\EE_{p_t}\norm{V_t}^2 \leq e^{-2t}\EE_p\norm{V}^2$.
\end{lemma}


\begin{proof}
By \Cref{lemma:realizable-field-equivalence}, we know that
\begin{equation} \EE_p[fh] = \EE_p\langle V,\grad h\rangle \qquad \forall h \in H^1(p),\label{eq:base-divergence-form}\end{equation}
and it suffices to check that 
\[\EE_{p_t}[f_th] = \EE_{p_t}\langle V_t, \grad h\rangle \qquad \forall h \in H^1(p_t).\]
Observe that by definition of $f_t$ and the Ornstein-Uhlenbeck process, for any $h \in H^1(p_t)$,
\begin{align*}
\EE_{p_t}[f_t h]
&= \EE[f_t(Z_t) h(Z_t)] \\ 
&= \EE[\EE[f(Z_0) \mid{} Z_t] h(Z_t)] \\ 
&= \EE[\EE[f(Z_0) h(Z_t) \mid{} Z_t]] \\ 
&= \EE[f(Z_0)h(Z_t)] \\ 
&= \EE_p[f \cdot (Q_th)]
\end{align*}
where $Q_t$ is the time-$t$ Ornstein Uhlenbeck operator
\[(Q_t h)(y) := \EE[h(Z_t) \mid{} Z_0 = y] = \EE_{\xi \sim \normal(0,I_d)}[h(e^{-t} y + \sqrt{1 - e^{-2t}} \xi)].\]
We can compute that
\[\grad (Q_t h)(y) = e^{-t}(Q_t \grad h)(y),\]
and by Jensen's inequality it holds that $Q_t h \in H^1(p)$. Applying \Cref{eq:base-divergence-form} to $Q_t h$, and then essentially reversing the algebraic manipulations above, gives
\begin{align*}
\EE_p[f \cdot (Q_t h)]
&= e^{-t}\EE_p \langle V, Q_t \grad h\rangle \\ 
&= e^{-t}\EE[\langle V(Z_0), \grad h(Z_t) \rangle] \\ 
&= e^{-t}\EE[\EE[\langle V(Z_0), \grad h(Z_t)\rangle \mid{} Z_t]] \\ 
&= \EE\langle V_t(Z_t), \grad h(Z_t)\rangle \\ 
&= \EE_{p_t} \langle V_t, \grad h\rangle.
\end{align*}
This shows that $V_t$ realizes perturbation of $p_t$ by $f_t$. The energy bound is immediate from Jensen's inequality.
\end{proof}

\subsection{Variance decay and the integrated Dirichlet energy}\label{sec:variance-decay}

Using the Poincar\'e-type bound from \Cref{lemma:sqrt-poincare}, we show that the variance of the posterior statistics decays exponentially, which enables a horizon-independent bound on the integrated Dirichlet energy \eqref{eq:dirichlet-integral-section-outline}.

\begin{lemma}\label{lemma:variance-decay}
Let $f: \RR^d \to \RR$ be a test statistic. For all $0 \leq t \leq T$, it holds that
\[\Var_{p_t}(f_t) \leq e^{-2t} \max(1,C_P(p;f)) \Var_p(f).\]
\end{lemma}

\begin{proof}
Combining \Cref{lemma:sqrt-poincare} with \eqref{eq:dissipation}, we get a differential inequality in $\Var_{p_t}(f_t)$:
\[\Var_{p_t}(f_t) \leq e^{-t} \sqrt{-\frac{1}{2}C_P(p;f)\Var_p(f) \frac{\dd}{\dd t} \Var_{p_t}(f_t)}.\]
Rearranging,
\[\frac{\dd}{\dd t} \Var_{p_t}(f_t) \leq -\frac{2e^{2t}}{C_P(p;f)\Var_p(f)} \Var_{p_t}(f_t)^2.\]
Thus, $\Var_{p_t}(f_t)$ is dominated by the function $g$ that solves the ODE
\[g'(t) = -\frac{2e^{2t}}{C_P(p;f)\Var_p(f)} g(t)^2\]
with $g(0) = \Var_p(f)$. One may check that the solution is defined by
\[\frac{1}{g(t)} = \frac{e^{2t}-1}{C_P(p;f)\Var_p(f)} + \frac{1}{\Var_p(f)},\]
and hence 
\[\Var_{p_t}(f_t) \leq g(t) \leq e^{-2t}\max(1,C_P(p;f))\Var_p(f)\]
as claimed.
\end{proof}

\begin{lemma}\label{lemma:dirichlet-integral-bound}
Let $f: \RR^d \to \RR$ be a test statistic. Write $C := \max(C_P(p;f), 1)$. Then
\[\int_0^T \sqrt{\EE_{p_t} \norm{\grad f_t}^2} \,\dd t \leq 2\sqrt{(1 + \log(C))\Var_p(f)}.\]
\end{lemma}

\begin{proof}
We can bound the integral by decomposing into appropriate intervals, applying Cauchy-Schwarz and \eqref{eq:dissipation} on each interval, and then using \Cref{lemma:variance-decay} for each interval. Set $a := \frac{1}{2}\log(C)$. Then,
\begin{align*}
\int_0^T \sqrt{\EE_{p_t} \norm{\grad f_t}^2} \, \dd t &\leq \int_0^a \sqrt{\EE_{p_t} \norm{\grad f_t}^2} \, \dd t + \sum_{n=0}^\infty \int_{a+n}^{a+n+1} \sqrt{\EE_{p_t} \norm{\grad f_t}^2} \, \dd t \\ 
&\leq \sqrt{a\int_0^a \EE_{p_t} \norm{\grad f_t}^2 \, \dd t} + \sum_{n=0}^\infty \sqrt{\int_{a+n}^{a+n+1} \EE_{p_t} \norm{\grad f_t}^2 \, \dd t} \\ 
&\leq \sqrt{a\Var_p(f)} + \sum_{n=0}^\infty \sqrt{\Var_{p_{a+n}}(f_{a+n}) - \Var_{p_{a+n+1}}(f_{a+n+1})} \\ 
&\leq \sqrt{a\Var_p(f)} + \sum_{n=0}^\infty e^{-a-n}\sqrt{\max(1,C_P(p;f))\Var_p(f)} \\ 
&\leq \left(\sqrt{\frac{1}{2}\log(C)} + \frac{e}{e-1}\right)\sqrt{\Var_p(f)}.
\end{align*}
This implies the claimed bound.
\end{proof}

\subsection{Proofs of main results for the learned reverse SDE}\label{sec:main-proofs}

The proofs of \Cref{thm:observable-main} and its corollaries are now straightforward. 

\begin{proof}[Proof of \Cref{thm:observable-main}]
First suppose that $\phat_0 = p_T$, i.e. the initialization of the learned reverse SDE is exact. Applying \Cref{lem:whole-comparison} to the (true and learned) reverse diffusion SDE, and invoking \Cref{assm:score-error}, we get
\[|\EE_{p}[f] - \EE_{\phat_T}[f]| \leq 2\epsscore \int_0^T \sqrt{\EE_{p_t}\norm{\grad f_t}^2} \, \dd t + \epsscore^2 T.\]
Applying \Cref{lemma:dirichlet-integral-bound} gives
\[|\EE_{p}[f] - \EE_{\phat_T}[f]| \leq 4\epsscore \sqrt{(1+\log(C))\Var_p(f)} + \epsscore^2 T.\]
Since $0 \leq f \leq 1$, we can bound the variance of $f$ by $1/4$.

Finally, if the initialization of the learned reverse SDE is drawn from some arbitrary distribution $\phat_0$ with $\eta_T = \TV(\phat_0, p_T)$, then the induced total variation error of the entire process is at most $\eta_T$ by the data processing inequality. Since $0 \leq f \leq 1$, this error contributes at most $\eta_T$ to the overall discrepancy.
\end{proof}

\begin{corollary}[Bounded error accumulation under Poincar\'e]\label{cor:poincare-main}
Write $C := \max(C_P(p), 1)$. The sampling law $\phat_T$, induced by the learned reverse SDE, satisfies
\begin{align}
 \TV(\phat_T,p)
 &\leq\eta_T+2\epsscore\sqrt{1+\log(C)}
       +\epsscore^2T.
 \label{eq:poincare-whole-bound}
\end{align}
\end{corollary}

\begin{proof}[Proof of \Cref{cor:poincare-main}]
By \Cref{lemma:local-global-equivalence}, we have $C_P(p;f) \leq C_P(p)$ (and hence $C_f := \max(C_P(p;f),1) \leq C$) for all bounded test statistics $f:\RR^d \to [0,1]$ with positive variance. Thus, for all such $f$, \Cref{thm:observable-main} gives
\[|\EE_{\phat_T}f-\EE_p f| \leq \eta_T + 2\epsscore \sqrt{1+\log(C)} + \epsscore^2T.\]
Taking the supremum over $f$ (and using an approximation argument to pass to non-smooth functionals) yields the claimed total variation bound.
\end{proof}

Below, suppose that $p$ is a mixture distribution where each component satisfies a Poincar\'e inequality:
\begin{align}
 p=\sum_{i=1}^m w_iq_i,\qquad \text{ where }\quad
 w_i>0,\quad \sum_{i=1}^m w_i=1,\quad C_P(q_i)\le K<\infty.
 \label{eq:mixture-assumption}
\end{align}
Then, \cref{thm:observable-main} applies so long as $f$ is \emph{balanced} in expectation across the different components:

\begin{lemma}\label{prop:mixture-flux}
In the setting of \eqref{eq:mixture-assumption}, let $f \in L^2(p)$ be a test statistic satisfying the balance condition $\EE_{q_1}[f] = \dots = \EE_{q_m}[f]$. Then $C_P(p;f) \leq K.$
\end{lemma}

\begin{proof}
Without loss of generality, we may assume that $\EE_p[f] = 0$, and hence $\EE_{q_i}[f] = 0$ for all $i \in [m]$. We may also assume that $w_i > 0$ for all $i \in [m]$.

We must show that $T_p(f) \leq K \Var_p(f)$. To do so, one can either construct a realizable field $V$ with bounded energy, or bound the dual formulation provided by \Cref{prop:flux-duality}; we choose the latter.

For any $h \in H^1(p)$, we can decompose 
\[\EE_p[fh] = \sum_{i=1}^m w_i \EE_{q_i}[fh] = \sum_{i=1}^m w_i \EE_{q_i}[f(h - \EE_{q_i}h)] \leq \sum_{i=1}^m w_i \sqrt{\Var_{q_i}(f)\Var_{q_i}(h)}\]
using the fact that $f$ is centered with respect to every $q_i$. Fix a parameter $\lambda>0$ to be determined later. By the AM-GM inequality,
we have for each $i \in [m]$ that
\[\sqrt{\Var_{q_i}(f)\Var_{q_i}(h)} \leq \frac{\lambda}{2} \Var_{q_i}(f) + \frac{1}{2\lambda}\Var_{q_i}(h).\]
Thus, summing over $i \in [m]$,
\begin{align*}
\EE_p[fh]
&\leq \frac{\lambda}{2}\sum_{i=1}^m w_i \Var_{q_i}(f) + \frac{1}{2\lambda} \sum_{i=1}^m w_i \Var_{q_i}(h) \\ 
&\leq \frac{\lambda}{2}\Var_p(f) + \frac{K}{2\lambda} \EE_p \norm{\grad h}^2
\end{align*}
where the second inequality uses the assumption that $f$ is balanced (and hence $\Var_p(f) = \sum_{i=1}^m w_i \Var_{q_i}(f)$) as well as the Poincar\'e bounds on the components $q_i$. Choosing $\lambda := \sqrt{\frac{K\EE_p\norm{\grad h}^2}{\Var_p(f)}}$ to optimize the bound, we get
\[\EE_p[fh] \leq \sqrt{K\Var_p(f)\EE_p\norm{\grad h}^2}.\]
Applying \Cref{prop:flux-duality}, we get $\TT_p(f) \leq K\Var_p(f)$, and hence $C_P(p;f) \leq K$.
\end{proof}

\begin{corollary}\label{cor:balanced-main}
In the setting of \eqref{eq:mixture-assumption}, let $f:\mathbb R^d\to[0,1]$ be a bounded test statistic for $p$, satisfying the \emph{balance condition}
$\EE_{q_1}f=\cdots=\EE_{q_m}f$. Then the sampling law $\phat_T$, induced by the learned reverse SDE, satisfies
\begin{align}
 |\EE_{\phat_T} f - \EE_p f|
 &\le\eta_T+2\epsscore\sqrt{1+\log\max(1,K)}
       +\epsscore^2T.
 \label{eq:balanced-whole-bound}
\end{align}
\end{corollary}

\begin{proof}[Proof of \Cref{cor:balanced-main}]
Immediate from \Cref{thm:observable-main} and \Cref{prop:mixture-flux}.
\end{proof}

\section{The information-theoretic lower bound}
\label{sec:it-lower}
In this section, we prove \Cref{prop:it-lower-intro}, formally restated below as \Cref{prop:it-lower-main}. It asserts that information-theoretically, given only a bound $K$ on the Poincar\'e constant of the data distribution, any sampler necessarily incurs $\Omega(\epsscore\sqrt{\log(K)})$ total variation error (unless $K$ is exponentially large in the dimension):

\begin{proposition}\label{prop:it-lower-main}
Let $d \in \NN$, $K \geq 1$, and $\epsscore>0$. There are distributions $p,q$ on $\RR^d$ and a function $\shat: \RR^d \times [0,T] \to \RR^d$ such that 
\[\sup_{t} \ScoreErr{p_t}{\shat_t} = \sup_{t} \ScoreErr{q_t}{\shat_t} \leq \epsscore^2\]
and $C_P(p) = C_P(q) \leq K$, but
\[\TV(p,q) \gtrsim \min\left\{1, \epsscore \sqrt{\min(d, \log K)}\right\}.\]
\end{proposition}

\begin{remark}[On worst-case tightness of the path-space analysis]\label{remark:worst-case-tightness}
The distributions $p,q$ constructed in the above proposition satisfy
\[\max\{\Dkl{p}{\normal(0,I_d)},\Dkl{q}{\normal(0,I_d)}\} \asymp \min(K, 2^d)\]
in the regime $\epsscore < 1$ and $K, d$ sufficiently large. Thus, with the optimal choice $T \asymp \min(d, \log K) + \log(1/\epsscore)$, the classical error bound \eqref{eq:standard-girsanov-analysis} shows that the diffusion sampler (with sufficiently fine discretization) recovers $p$ from estimated scores for $p$ with total variation error $O(\epsscore\sqrt{\min(d,\log K)} + \epsscore\sqrt{\log(1/\epsscore)})$ (and similarly for $q$). \Cref{prop:it-lower-main} shows that this error bound is optimal up to the additional term $\epsscore\log(1/\epsscore)$. In other words, for this particular example and this particular choice of $T$, the error accumulation suggested by the path-space analysis \eqref{eq:standard-girsanov-analysis} is (roughly) tight: the error does indeed accumulate like $\sqrt{T}$. However, this accumulation does not continue if $T$ is made still larger; also, it is straightforward to construct examples where the path-space analysis is loose (compared to \Cref{prop:it-upper-intro} or even \Cref{cor:poincare-intro}) even when $T$ is chosen optimally. 
\end{remark}

\paragraph{Construction: multi-scale Gaussians.} Fix a dimension $d \in \NN$ and a desired Poincar\'e bound $K \geq 1$. We will define distributions $p,q$ on $\RR^d$ to be multivariate Gaussian distributions with the same covariance but different mean. Concretely, we set $p := \normal(\mu, \Sigma)$ and $q := \normal(-\mu,\Sigma)$ where $\mu$ and $\Sigma$ are defined below. 

Let
\begin{equation}
 m:=\min\left\{d,1+\lfloor\log_2K\rfloor\right\},
 \qquad \lambda_j:=2^{j-1}\quad(1\le j\le m),
 \label{eq:lower-m-lambda}
\end{equation}
and define the shared covariance matrix by
\begin{equation}
 \Sigma:=\diag(\lambda_1,\ldots,\lambda_m,1,\ldots,1).
 \label{eq:lower-sigma}
\end{equation}
Define the mean vector $\mu$ by
\begin{equation}
 \mu_j:=\frac\epsscore2\sqrt{\lambda_j}\quad(1\le j\le m),
 \qquad \mu_j:=0\quad(j>m).
 \label{eq:lower-mu}
\end{equation}
We define the approximate score function to be
\[\shat(x,t) := \grad \log r_t(x)\]
where $r := \normal(0,\Sigma)$, and $r_t$ is the pushforward of $r$ under the time-$t$ Ornstein-Uhlenbeck process.

\paragraph{Analysis.} We bound  (1) the Poincar\'e constants of $p, q$; (2) the score error of $\shat$ with respect to each distributions; and (3) the total variation distance between $p$ and $q$.

\begin{claim}\label{claim:lower-poincare}
It holds that $C_P(p) = C_P(q) \leq K$.
\end{claim}

\begin{proof}
The Poincar\'e constant of a multivariate Gaussian is equal to the largest eigenvalue of its covariance matrix. By definition of $m$, we have $\lambda_m = 2^{m-1} \leq K$.
\end{proof}

\begin{claim}
\label{claim:lower-score}
It holds that
\begin{equation}
 \sup_{t\ge0}\ScoreErr{p_t}{\shat_t}
 =\sup_{t\ge0}\ScoreErr{q_t}{\shat_t}
 \le\epsscore^2.
 \label{eq:lower-score-bound}
\end{equation}
\end{claim}

\begin{proof}
Fix $t \geq 0$. We bound $\ScoreErr{p_t}{\shat_t}$; the argument for $q_t$ is identical. Observe that $p_t = \normal(\mu_t, \Sigma_t)$ and $r_t = \normal(0, \Sigma_t)$ where
\[\mu_t := e^{-t} \mu = \frac{e^{-t}\epsscore}{2}(\sqrt{\lambda_1},\dots,\sqrt{\lambda_m},0,\dots,0)\]
and
\[\Sigma_t := e^{-2t}\Sigma + (1-e^{-2t})I_d = \diag\left(e^{-2t}\lambda_1+1-e^{-2t},\dots, e^{-2t}\lambda_m+1-e^{-2t},1,\dots,1\right).\]
Since $\grad \log p_t(x) = -\Sigma_t^{-1}(x - \mu_t)$ and $\grad \log r_t(x) = -\Sigma_t^{-1}x$, we have
\begin{align*}
\ScoreErr{p_t}{\shat_t}
&= \EE_{x\sim p_t} \norm{\grad \log p_t(x) - \grad \log r_t(x)}^2 \\ 
&= \mu_t \Sigma_t^{-2} \mu_t \\
&= \frac{e^{-2t}\epsscore^2}{4}\sum_{j=1}^m \frac{\lambda_j}{(e^{-2t}\lambda_j + 1 - e^{-2t})^2}. 
\end{align*}
Choose $j_0 \in \{0,\dots,m\}$ so that $\lambda_j \leq e^{2t}$ for $j \leq j_0$ and $\lambda_j > e^{2t}$ for $j > j_0$. Since the denominator in the summand is always at least $1$, and $(\lambda_j)_j$ is a dyadic exponential sequence, we can bound
\[\sum_{j=1}^{j_0} \frac{\lambda_j}{(e^{-2t}\lambda_j + 1 - e^{-2t})^2} \leq \sum_{j=1}^{j_0} \lambda_j \leq 2\lambda_{j_0} \leq 2e^{2t}.\]
Since the denominator is always at least $e^{-4t}\lambda_j^2$, we can bound
\[\sum_{j=j_0+1}^{m} \frac{\lambda_j}{(e^{-2t}\lambda_j + 1 - e^{-2t})^2} \leq \sum_{j=j_0+1}^m \frac{e^{4t}}{\lambda_j} \leq \frac{2e^{4t}}{\lambda_{j_0+1}} \leq 2e^{2t}.\]
Putting together these bounds gives
\[\ScoreErr{p_t}{\shat_t} \leq \epsscore^2\]
as claimed.
\end{proof}

\begin{claim}
\label{claim:lower-tv}
It holds that
\[\TV(p,q)
 \geq c\min\{1,\eps\sqrt m\}
 \label{eq:lower-tv}
\]
for an absolute constant $c>0$.
\end{claim}

\begin{proof}
Apply the invertible linear map $x\mapsto\Sigma^{-1/2}x$.  The two laws become $\mathcal N(v,I_d)$ and $\mathcal N(-v,I_d)$ with
\begin{equation}
 \norm v^2=\mu^\top\Sigma^{-1}\mu
 =\sum_{j=1}^m\frac{\mu_j^2}{\lambda_j}
 =\frac{\eps^2m}{4}.
 \label{eq:lower-vnorm}
\end{equation}
By data processing, we have 
\[\TV(p,q) \geq \TV(\normal(\norm{v}, 1), \normal(-\norm{v}, 1)).\]
The claim then follows from standard results, e.g. \cite[Theorem 3.1]{li2019totalvariation}.
\end{proof}

\begin{proof}[Proof of \Cref{prop:it-lower-main}]
Immediate from \Cref{claim:lower-poincare,claim:lower-score,claim:lower-tv}.
\end{proof}

\section{The information-theoretically optimal algorithm}
\label{sec:projection}

In this section, we prove \Cref{prop:it-upper-intro}, formally restated below as \Cref{prop:it-upper-main}, which asserts that an annealed score with $L^2$ error $\epsscore^2$ information-theoretically identifies the distribution up to total variation error $O(\epsscore\sqrt{1+\log(K)})$, where $K\geq 1$ bounds the Poincar\'e constant of the distribution. 

\begin{proposition}\label{prop:it-upper-main}
Let $d \in \NN$, $K \geq 1$, and $\epsscore > 0$. There is a map $\cA$ with the following property.

Let $p$ be a distribution on $\RR^d$ and let $\shat: \RR^d \times [0,\infty) \to \RR^d$ be a function. If $C_P(p) \leq K$ and
\[\sup_{t \geq 0} \ScoreErr{p_t}{\shat_t} \leq \epsscore^2,\]
then $\TV(p, \cA(\shat)) \lesssim \epsscore \sqrt{1+\log(K)}$.
\end{proposition}

We first observe that it is an immediate consequence of the following lemma; the rest of the section is devoted to the proof of the lemma.

\begin{lemma}[Two-point total variation bound under Poincar\'e]\label{lemma:two-point}
Fix $K \geq 1$. Let $p,q$ be distributions on $\RR^d$ with $C_P(p) \leq K$ and $C_P(q) \leq K$. Let $\shat: \RR^d \times [0,\infty) \to \RR^d$ be a function satisfying
\[\sup_{t \geq 0} \max\{\ScoreErr{p_t}{\shat_t}, \ScoreErr{q_t}{\shat_t}\} \leq \epsscore^2.\]
Then $\TV(p,q) \leq \epsscore(16+\sqrt{\log(K)})$.
\end{lemma}

\begin{proof}[Proof of \Cref{prop:it-upper-main}]
Define the map $\cA$ as follows. Given $\shat$, let $\cA(\shat)$ denote any $q$ that satisfies $C_P(q) \leq K$ and $\sup_{t \geq 0} \ScoreErr{q_t}{\shat_t} \leq \epsscore^2$.

By assumption, $p$ satisfies the same Poincar\'e bound and score error bound. Thus, \Cref{lemma:two-point} gives $\TV(p,q) \leq \epsscore(16+ \sqrt{\log(K)})$, which implies the claimed bound. 
\end{proof}

As discussed in \Cref{sec:overview-it}, the proof of \Cref{lemma:two-point} starts with a Girsanov-based analysis.

\begin{lemma}\label{lem:final-girsanov}
Let $p,q$ be distributions and let $\tstar > 0$. Let $\shat: \RR^d \times [0,\infty) \to \RR^d$ be a function satisfying
\[\sup_{t \geq 0} \max\{\ScoreErr{p_t}{\shat_t}, \ScoreErr{q_t}{\shat_t}\} \leq \epsscore^2.\]
Then \[\TV(p,q) \leq \TV(p_{\tstar},q_{\tstar}) + \epsscore \sqrt{2\tstar}.\]
\end{lemma}

\begin{proof}
Let $P$ be the path law of the reverse Ornstein-Uhlenbeck diffusion with initialization $p_{\tstar}$ at time $\tstar$, and with the true annealed score functions $\grad \log p_t$. Let $\wh P$ be the path law with initialization $p_{\tstar}$ but with the estimated score functions $\shat_t$. Similarly define $Q$ and $\wh Q$. By Girsanov's theorem (as applied in \cite{chen2023sampling}) and Pinsker's inequality,
\[\TV(P,\wh P) \leq \sqrt{\frac{1}{2}\Dkl{P}{\wh P}} \leq \frac{1}{\sqrt{2}}\epsscore\sqrt{\tstar}.\]
Similarly, $\TV(Q,\wh Q) \leq \frac{1}{\sqrt{2}}\epsscore\sqrt{\tstar}$. Finally, since the processes $\wh P$ and $\wh Q$ have identical conditional laws for fixed initialization, the data processing inequality gives $\TV(\wh P,\wh Q) \leq \TV(p_{\tstar},q_{\tstar})$. It follows from the triangle inequality that
\[\TV(P, Q) \leq \TV(p_{\tstar},q_{\tstar}) + \epsscore\sqrt{2\tstar}.\]
By construction, the final marginal law of $P$ is $p$, and the final marginal law of $Q$ is $q$. Thus, applying the data processing inequality completes the proof.
\end{proof}

It remains to show that $\TV(p_{\tstar},q_{\tstar}) \leq O(\epsscore)$ with $\tstar := \frac{1}{2}\log(K)$. Towards this end, we require several preparatory lemmas. First, note that the estimated score in \Cref{lemma:two-point}, at any time $t$, is not required to be the score of a real distribution. The following lemma implies that we may assume without loss of generality that it is --- and in fact, we may assume that it is the score of the \emph{mixture distribution}.

\begin{lemma}
\label{lem:mixture-score}
Let $\mu,\nu$ be distributions on $\RR^d$ and let $\shat:\RR^d \to \RR^d$ be a function. Then
\begin{equation}
 \ScoreErr{\mu}{\grad \log r}+\ScoreErr{\nu}{\grad \log r}
 \leq \ScoreErr{\mu}{\shat}+\ScoreErr{\nu}{\shat}
 \label{eq:mixture-score-opt}
\end{equation}
where $r := \frac{\mu+\nu}{2}$.
\end{lemma}

\begin{proof}
For any fixed $x \in \RR^d$, the function
\[
z \mapsto \mu(x)\norm{z-\grad \log \mu(x)}^2+\nu(x)\norm{z-\grad \log \nu(x)}^2
\]
is minimized at
\[
 z := \frac{\mu(x)\grad \log \mu(x)+\nu(x)\grad \log \nu(x)}{\mu(x)+\nu(x)}
 =\frac{\grad \mu(x)+\grad \nu(x)}{\mu(x)+\nu(x)}
 =\grad \log r(x).
\]
Thus,
\begin{align*} 
&\mu(x)\norm{\grad \log r(x)-\grad \log \mu(x)}^2+\nu(x)\norm{\grad \log r(x)-\grad \log \nu(x)}^2 \\ 
&\leq \mu(x)\norm{\shat(x)-\grad \log \mu(x)}^2+\nu(x)\norm{\shat(x)-\grad \log \nu(x)}^2.
\end{align*}
Integrating over $x\in \RR^d$ yields the claimed inequality.
\end{proof}

This enables the strategy of showing that $p_t$ and $q_t$  are close (at the critical time $t=\tcrit$; see \Cref{sec:overview-it}) by interpolating through $r := \frac{p_t+q_t}{2}$. To make this work, we need to control the Poincar\'e constant of $r$, and then use this to transfer from $\ScoreErr{p_t}{\grad \log r}$ to $\TV(p_t,r)$ (and similarly for $q_t$). The following standard lemma shows that the Poincar\'e constant of a mixture can be bounded by the Poincar\'e constants of its components, so long as the components have non-trivial overlap.\footnote{See e.g. \cite[Theorem 4]{koehler2024efficiently} for a version with more than two components.}

\begin{lemma}
\label{lem:mixture-poincare}
Let $\mu,\nu$ be distributions on $\RR^d$ and let $r=(\mu+\nu)/2$. If $\TV(\mu,\nu) \leq 1/2$, then
\begin{equation}
 C_P(r)\le3\max\{C_P(\mu),C_P(\nu)\}.
 \label{eq:mixture-poincare}
\end{equation}
\end{lemma}

\begin{proof}
Define $\alpha := \mu - \min(\mu,\nu)$, $\beta := \nu - \min(\mu,\nu)$, and $\gamma := \min(\mu,\nu)$. By assumption, $\int \gamma(x) \, dx = 1-\TV(\mu,\nu) \geq 1/2$. Define $\ol \gamma := \frac{\gamma}{\int \gamma(x) \, dx}$. For any smooth function $f:\RR^d \to \RR$, we have
\begin{align*} 
\Var_\mu(f) 
&= \EE_\mu(f - \EE_\mu f)^2 \\ 
&\geq \int \gamma(x) (f(x) - \EE_\mu f)^2 \, dx \\ 
&= \int \gamma(x) \, dx \cdot \EE_{\ol\gamma} (f - \EE_\mu f)^2 \\ 
&\geq \int \gamma(x) \, dx \cdot (\EE_{\ol\gamma} f - \EE_\mu f)^2.
\end{align*}
It follows that $(\EE_{\ol\gamma} f - \EE_\mu f)^2 \leq 2\Var_\mu(f)$. Similarly, $(\EE_{\ol\gamma} f - \EE_\nu f)^2 \leq 2\Var_\nu(f)$. We conclude that
\begin{align*}
\Var_r(f) 
&= \frac{1}{2}\Var_\mu(f) + \frac{1}{2}\Var_\nu(f) + \frac{1}{4}(\EE_\mu f - \EE_\nu f)^2  \\ 
&\leq \frac{1}{2}\Var_\mu(f) + \frac{1}{2}\Var_\nu(f) + \frac{1}{2} (\EE_\mu f - \EE_{\ol\gamma} f)^2 + \frac{1}{2} (\EE_{\ol \gamma} f - \EE_\nu f)^2 \\ 
&\leq \frac{3}{2}\Var_\mu(f) + \frac{3}{2} \Var_\nu(f) \\ 
&\leq \frac{3}{2} C_P(\mu) \EE_\mu \norm{\grad f}_2^2 + \frac{3}{2} C_P(\nu) \EE_\nu \norm{\grad f}_2^2 \\ 
&\leq 3 \max(C_P(\mu), C_P(\nu)) \cdot \EE_r \norm{\grad f}_2^2
\end{align*}
as needed.
\end{proof}

Finally, we show that a score error bound transfers to a $\chi^2$ bound under a Poincar\'e inequality on the distribution corresponding to the ``estimated score.'' This result can be thought of as an analogue of a standard characterization of the log-Sobolev constant \cite[Definition 2]{koehler2022statistical}.

\begin{lemma}
\label{lem:fisher-chi}
Let $\mu, r$ be distributions on $\RR^d$ with $\norm{\mu/r}_{L^\infty(r)} < \infty$. Then,
\begin{equation}
 \Dchi{\mu}{r}
 \le C_P(r)\norm{\mu/r}_{L^\infty(r)}
 \ScoreErr{\mu}{\grad \log r}.
 \label{eq:fisher-chi}
\end{equation}
\end{lemma}

\begin{proof}
We have
\begin{align*}
\Dchi{\mu}{r}
&= \EE_{x\sim r} \left(\frac{\mu(x)}{r(x)} - 1\right)^2 \\ 
&\leq C_P(r) \cdot \EE_r \norm{ \grad \frac{\mu(x)}{r(x)}}_2^2 \\ 
&= C_P(r) \cdot \EE_r\left[ \frac{\mu(x)^2}{r(x)^2}\cdot \norm{\grad \log \frac{\mu(x)}{r(x)}}_2^2\right] \\ 
&= C_P(r) \cdot \EE_\mu\left[ \frac{\mu(x)}{r(x)}\cdot \norm{\grad \log \frac{\mu(x)}{r(x)}}_2^2\right] \\ 
&\leq C_P(r) \norm{\frac{\mu}{r}}_\infty \cdot \ScoreErr{\mu}{\grad \log r}
\end{align*}
as needed.
\end{proof}

Using the above preparatory results, we can now bound the initialization error at $\tstar := \frac{1}{2}\log(K)$.

\begin{lemma}[Initialization error]
\label{prop:coarse-closeness}
Fix $K \geq 1$. Let $p,q$ be distributions on $\RR^d$ with $C_P(p) \leq K$ and $C_P(q) \leq K$. Let $\shat: \RR^d \times (0,1) \to \RR^d$ be a function satisfying
\begin{equation}\sup_{t \geq 0} \max\{\ScoreErr{p_t}{\shat_t}, \ScoreErr{q_t}{\shat_t}\} \leq \epsscore^2.\label{eq:score-error-bound}
\end{equation}
Then $\TV(p_{\tstar},q_{\tstar}) \leq 16\epsscore$ where $\tstar := \frac{1}{2}\log K$.
\end{lemma}

\begin{proof}
Define the \emph{critical time}
\[\tcrit := \inf\{t \geq \tstar: \TV(p_t, q_t) \leq 8\epsscore\}.\]
We prove that $\tcrit = \tstar$ by contradiction. Indeed, assume $\tcrit > \tstar$. Then $\TV(p_{\tstar},q_{\tstar}) > 8\epsscore$. By continuity of the map $t \mapsto \TV(p_t,q_t)$ and the fact that it converges to $0$ as $t \to \infty$ (\Cref{lemma:tv-continuity-convergence}), we have that $\tcrit < \infty$ and $\TV(p_{\tcrit},q_{\tcrit}) = 8\epsscore$.

We may assume that $\epsscore < 1/16$ (otherwise, the lemma's claim is vacuously true), so $\TV(p_{\tcrit},q_{\tcrit}) < 1/2$. By \Cref{lem:poincare-decay} and the fact that $\tcrit \geq \frac{1}{2}\log K$, we have $C_P(p_{\tcrit}) \leq 2$ and $C_P(q_{\tcrit}) \leq 2$. Thus, $C_P(r) \leq 6$ where $r := \frac{p_{\tcrit}+q_{\tcrit}}{2}$. 

By \Cref{lem:mixture-score} and \eqref{eq:score-error-bound}, we have
\begin{align*} 
&\ScoreErr{p_{\tcrit}}{\grad \log r} + \ScoreErr{q_{\tcrit}}{\grad \log r} 
\\
&\leq \ScoreErr{p_{\tcrit}}{\shat_{\tcrit}} + \ScoreErr{q_{\tcrit}}{\shat_{\tcrit}}  \leq 2\epsscore^2.
\end{align*}
Thus, by \Cref{lem:fisher-chi}, we have
\[\TV(p_{\tcrit}, r) \leq \sqrt{\frac{1}{2}\Dchi{p_{\tcrit}}{r}} \leq \sqrt{12\epsscore^2} < 4\epsscore\]
and similarly $\TV(q_{\tcrit},r) < 4\epsscore$. It follows that $\TV(p_{\tcrit},q_{\tcrit}) < 8\epsscore$, which is a contradiction. Thus, in fact $\tcrit = \tstar$, which concludes the proof.
\end{proof}

\begin{proof}[Proof of \Cref{lemma:two-point}]
Immediate from \Cref{lem:final-girsanov} and \Cref{prop:coarse-closeness}.
\end{proof}

\section{Tighter bounds via time-to-isoperimetry}

In this section, we show that the arguments behind our main results on the error of the learned reverse SDE actually extend to a broader family of data distributions $p$---including distributions with bounded support yet arbitrarily bad Poincar\'e constant. For simplicity, we focus on the goal of bounding total variation error, though the same idea works for individual test statistics.

The basic intuition behind all of our main results is that functional inequalities exponentially improve under the Ornstein-Uhlenbeck process. For instance, we know that if $p$ has Poincar\'e constant $C_P(p) \leq K$, then $p_t$ has Poincar\'e constant at most $e^{-2t} K + 1-e^{-2t}$ (\Cref{lem:poincare-decay}). The $\log(C_P(p))$ dependence in \Cref{cor:poincare-intro} then essentially comes from the fact that $C_P(p_t) \leq O(1)$ for all $t \geq \frac{1}{2}\log(K)$. Heuristically, this isoperimetry suggests that errors in the learned reverse SDE should ``self-correct'' on the interval $[\frac{1}{2}\log(K), T]$, and thus errors only accumulate on the interval $[0, \frac{1}{2}\log(K)]$.

This intuition can be made explicit in the following sense:

\begin{theorem}\label{thm:time-to-poincare}
The sampling law $\phat_T$, induced by the learned reverse SDE, satisfies
\[\TV(\phat_T, p) \lesssim \eta_T + \epsscore\sqrt{1+\tstar} + \epsscore^2 T\]
where $\tstar := \min\{t\geq 0: C_P(p_t) \leq 2\}$ is the \emph{time-to-isoperimetry} for $p$ (under the Ornstein-Uhlenbeck diffusion).
\end{theorem}

It always holds that $\tstar \leq \frac{1}{2}\log(\max(1,C_P(p)))$, and thus \Cref{thm:time-to-poincare} generalizes \Cref{cor:poincare-main} (the formal version of \Cref{cor:poincare-intro}), but in certain cases $\tstar$ can be substantially smaller. In particular, suppose that $p$ is known to be supported on the Euclidean ball of radius $R \geq 1$ (this violates the regularity conditions from \Cref{sec:conventions}, but it will not matter). Generically, $p$ can have arbitrarily large Poincar\'e constant; nevertheless, the smoothing property of Ornstein-Uhlenbeck diffusion ensures that the time-to-isoperimetry is at most $1+\log(R)$ (\Cref{lem:bounded-radius-poincare}). Thus, we get the following immediate corollary:

\begin{corollary}\label{cor:bounded-support}
Suppose that $p$ is supported on a Euclidean ball of radius $R \geq 1$. Then the sampling law $\phat_T$, induced by the learned reverse SDE, satisfies
\[\TV(\phat_T, p) \lesssim \eta_T + \epsscore\sqrt{1+\log(R)} + \epsscore^2 T.\]
\end{corollary}

We briefly sketch the proof of \Cref{thm:time-to-poincare}, which can be derived in a black-box fashion from \Cref{cor:poincare-main}.

\begin{proof}[Proof of \Cref{thm:time-to-poincare}]
Assume that the learned reverse SDE is initialized at $p_T$; we can remove this assumption at the end, at the cost of the extra error term $\eta_T = \TV(p_T, \phat_0)$.

Without loss of generality suppose $\tstar > 0$ (otherwise we can apply the following argument to $\tstar+1$). If $\tstar > T$ then we can apply the standard path-space analysis (Pinsker's inequality together with Girsanov's theorem, which is applicable by the argument in \cite{chen2023sampling}). Thus, we may assume $\tstar \in (0,T)$. Apply \Cref{cor:poincare-main} to $p_{\tstar}$ (which satisfies the regularity conditions from \Cref{sec:conventions} since $\tstar>0$). Since $C_P(p_{\tstar}) \leq 2$, this shows that the total variation error incurred by the learned reverse SDE from time $T$ down to time $\tstar$ is at most $O(\epsscore+\epsscore^2 T)$. By the data processing inequality, this bounds the error between $\phat_T$ and the terminal marginal of the reverse SDE on $[0,T]$ that uses the learned score on $[0,\tstar)$ and the true score on $[\tstar, T]$. The error between this marginal and $p$ is at most $\epsscore\sqrt{\tstar}$ by the standard path-space analysis. Combining these error bounds and adding in the initialization error yields the claimed result.
\end{proof}

\begin{remark}\label{remark:bounded-support}
By contraction of the Ornstein-Uhlenbeck operator, we can bound the initialization error by $\eta_T \leq e^{1-T}\sqrt{\Dkl{p_1}{\normal(0,I_d)}}$. If $p$ is supported on the Euclidean ball of radius $R$, then $\Dkl{p_1}{\normal(0,I_d)} \lesssim R^2+d$ by \Cref{lemma:kl-ub}. Substituting into \Cref{cor:bounded-support} and optimizing the choice of $T$, we get
\[\TV(\phat_T,p) \lesssim \epsscore\sqrt{1+\log(R)} + \epsscore^2 \log((d+R)/\epsscore).\]
In contrast, the standard path-space analysis (with the same initialization error bound) gives
\begin{align*} 
\TV(\phat_T, p) 
&\lesssim \eta_T + \epsscore\sqrt{T} \\ 
&\lesssim e^{1-T} \sqrt{R^2 + d} + \epsscore\sqrt{T},
\end{align*}
which for the optimal choice of $T$ only yields
\[\TV(\phat_T,p) \lesssim \epsscore \sqrt{\log((d+R)/\epsscore)}.\]
Thus, \Cref{cor:bounded-support} is tighter in some regimes: for instance, if $R = O(1)$ and $\epsscore \leq 1/\log(d/\epsscore)$, then we get $\TV(\phat_T,p) \lesssim \epsscore$ rather than $\TV(\phat_T,p) \lesssim \epsscore\sqrt{\log(d/\epsscore)}$.
\end{remark}

\section{Technical results for the \spectralineq{f}}\label{sec:geometry}
In this section we prove several useful facts relating to the \spectralineq{f}: a dual condition for a vector field $V$ to realize a perturbation (\Cref{lemma:realizable-field-equivalence}), which we used to prove a dual expression for the transport energy $\TT_\mu(g)$ (\Cref{prop:flux-duality}; c.f. \eqref{eq:tangent-norm-definitions}); and finally an equivalence between the standard Poincar\'e inequality and the worst-case \spectralineq{f} over all test statistics $f$ (\Cref{lemma:local-global-equivalence}). The first two results are essentially standard technical results, which we prove for completeness.

\begin{lemma}\label{lemma:realizable-field-equivalence}
For every distribution $\mu$ on $\RR^d$, test statistic $g \in L^2(\mu)$, and vector field $V \in L^2(\mu;\RR^d)$, the following are equivalent:
\begin{enumerate}
    \item $V$ realizes perturbation of $\mu$ by $g$, i.e. $-\grad \cdot (\mu V) = g\mu$.
    \item $\EE_\mu[gh] = \EE_\mu \langle V, \grad h\rangle$ for all $h \in H^1(\mu)$.
\end{enumerate}
\end{lemma}

\begin{proof}
Suppose (1) holds. Pick any smooth $h \in H^1(\mu)$ and define $F := \mu h V$. Then $F$ is continuously differentiable (by assumptions in \Cref{sec:conventions}) and absolutely integrable (by the lemma assumptions). Moreover, $\grad \cdot F = -\mu gh + \mu\langle V,\grad h\rangle$ by (1), and hence is absolutely integrable. Thus, the divergence theorem (\Cref{lemma:divergence-theorem}) yields 
\[0 = \int_{\RR^d} \grad \cdot F(x) \,\dd x = -\EE_\mu[gh] + \EE_\mu \langle V,\grad h\rangle\]
which proves (2) for smooth $h$, and can be extended to all $h \in H^1(\mu)$ by limiting approximation. Conversely, suppose (2) holds. Then the function $f := g\mu + \grad \cdot (\mu V)$ satisfies
\[\int_{\RR^d} f(x) h(x) = 0\]
for all $h \in H^1(\mu)$. Since $f$ is continuous (see \Cref{sec:conventions}), it follows that $f$ is identically zero (\Cref{lemma:zero-integral}), which proves (1).
\end{proof}

\begin{lemma}[Primal--dual identity]\label{prop:flux-duality}
For every distribution $\mu$ on $\RR^d$ and centered test statistic $g\in L^2(\mu)$,
\begin{align}
 \mathsf T_\mu(g)
 &=\sup_{h\in H^1(\mu): \EE_\mu[\norm{\grad h}^2]>0} \frac{\EE_\mu[gh]^2}{\EE_\mu[\norm{\grad h}^2]}
 \label{eq:flux-duality}
\end{align}
\end{lemma}

\begin{proof}
Fix $h\in H^1(\mu)$. For any vector field $V$ that realizes perturbation of $\mu$ by $g$, we have
\begin{align*}
\EE_\mu[gh]^2
&=(\EE_\mu \langle V,\grad h\rangle)^2 \\ 
&\leq (\EE_\mu \norm{V}\norm{\grad h})^2 \\ 
&\leq \EE_\mu[\norm{V}^2]\EE_\mu[\norm{\grad h}^2].
\end{align*}
Taking the infimum over $V$ yields
\[\EE_\mu[gh]^2 \leq \TT_\mu(g) \EE_\mu[\norm{\grad h}^2].\]
This establishes the inequality 
\[ \mathsf T_\mu(g)
 \geq \sup_{h\in H^1(\mu): \EE_\mu[\norm{\grad h}^2]>0} \frac{\EE_\mu[gh]^2}{\EE_\mu[\norm{\grad h}^2]} =: S.\]
If $S = \infty$ then the equality holds, so we may henceforth assume that $S<\infty$. Then
$|\EE_\mu[gh]|\leq\sqrt{S\EE_\mu[\norm{\grad h}^2]}$ for all $h \in H^1(\mu)$ with $\EE_\mu[\norm{\grad h}^2]>0$, and by limiting approximation, the bound extends to all $h \in H^1(\mu)$. It follows for any $h_1, h_2 \in H^1(\mu)$ with $\grad h_1 = \grad h_2$ that $\EE_\mu[gh_1] = \EE_\mu[gh_2]$. 

For notational convenience, let $H := L^2(\mu;\RR^d)$, which denotes the Hilbert space on vector fields $V: \RR^d \to \RR^d$ with inner product $\langle V, W\rangle_H := \EE_\mu\langle V,W\rangle$. Let $G := \{\grad h: h \in H^1(\mu)\} \subseteq H$. By the above argument, there is a well-defined map $\tau: G \to \RR$ with $\tau(\grad h) = \EE_\mu[gh]$ for all $h \in H^1(\mu)$. Moreover, $\tau$ is bounded and linear. Thus, it extends to a bounded linear functional $\ol\tau$ on the closure $\ol G^H$, which is itself a Hilbert space. By the Riesz representation theorem, there is some $\Vstar \in \ol G^H$ with $\ol\tau(W) = \langle W, \Vstar\rangle_H$ for all $W \in \ol G^H$. Moreover, $\norm{\Vstar}_H = \norm{\ol\tau}_{H^\star} = \norm{\tau}_{H^\star}$.

It remains to rewrite these facts in the original notation. The former gives that
\[\EE_\mu[gh] = \tau(\grad h) = \EE_\mu \langle \grad h, \Vstar\rangle\]
for all $h \in H^1(\mu)$, so by \Cref{lemma:realizable-field-equivalence}, $\Vstar$ realizes perturbation of $\mu$ by $g$.\footnote{Technically, since $\Vstar$ is in the closure $\ol G^H$, it may not satisfy the necessary differentiability condition for the realizability condition to be well-defined; this can be fixed by working entirely with the ``weak'' definition given in \Cref{lemma:realizable-field-equivalence}.} The latter gives \[\sqrt{\EE_\mu \norm{\Vstar}^2} = \sup\{|\EE_\mu[gh]|: h\in H^1(\mu), \EE_\mu\norm{\grad h}^2 \leq 1\} = \sqrt{S}.\]
Together, we conclude that $\TT_\mu(g) \leq S$, which completes the proof.
\end{proof}

\begin{lemma}[Global Poincar\'e is equivalent to worst-case \spectralineq{f}]\label{lemma:local-global-equivalence}
For any distribution $\mu$ on $\RR^d$, it holds that
\[C_P(\mu) = \sup_{f \in L^2(\mu): \Var_\mu(f)>0} C_P(\mu;f).\]
\end{lemma}
\begin{proof}
Let $f \in L^2(\mu)$ with $\Var_\mu(f)>0$, and write $g = f - \EE_\mu f$. Then for any $h\in H^1(\mu)$ with $\EE_\mu h = 0$,
\[\EE_\mu[gh]^2 \leq \Var_\mu(f)\Var_\mu(h) \leq C_P(\mu) \Var_\mu(f) \EE_\mu[\norm{\grad h}^2]\]
by Cauchy-Schwarz and the definition of $C_P(\mu)$. Moreover, both sides are invariant to a constant shift in $h$, so the restriction $\EE_\mu h = 0$ can be removed. Thus, $\TT_\mu(g) \leq C_P(\mu)\Var_\mu(f)$, so $C_P(\mu;f) \leq C_P(\mu)$.

Now pick any $f \in H^1(\mu)$ with $\EE_\mu[\norm{\grad f}^2]>0$ and $\EE_\mu f = 0$, and write $K := \Var_\mu(f)/\EE_\mu[\norm{\grad f}^2]$. We claim that $C_P(\mu;f) \geq K$. Indeed,
\[T_\mu(f) \geq \frac{\EE_\mu[f^2]^2}{\EE_\mu[\norm{\grad f}^2]} = \frac{\Var_\mu(f)^2}{\EE_\mu[\norm{\grad f}^2]} = K \Var_\mu(f).\]
It follows that
\[\sup_{f \in L^2(\mu): \Var_\mu(f)>0} C_P(\mu;f) \geq \sup_{f \in H^1(\mu): \EE_\mu[\norm{\grad f}^2]>0} \frac{\Var_\mu(f)}{\EE_\mu[\norm{\grad f}^2]} = C_P(\mu).\]
This completes the proof.
\end{proof}

\section{Technical lemmas}

\subsection{Measure theory}

\begin{lemma}[Divergence theorem]\label{lemma:divergence-theorem}
Let $F: \RR^d \to \RR^d$ be a continuously differentiable vector field. If $F$ and $\grad \cdot F$ are absolutely integrable, then
\[\int_{\RR^d} \grad \cdot F(x) \, \dd x = 0.\]
\end{lemma}

\begin{lemma}\label{lemma:zero-integral}
Let $\mu$ be a distribution on $\RR^d$. Any continuous function $f: \RR^d \to \RR$ satisfying
\[\int_{\RR^d} f(x)h(x) \, \dd x = 0 \qquad \forall h \in H^1(\mu)\]
must be identically zero.
\end{lemma}

\begin{proof}
Suppose there exists $x$ with $f(x)>0$. Then by continuity, $f(y)>0$ for all $y$ in an open ball around $x$. Let $h$ be a smooth bump function supported in this ball; since it is smooth and bounded, necessarily it lies in $H^1(\mu)$. However, clearly $\int_{\RR^d} f(x)h(x) \, \dd x > 0$. Thus $f \leq 0$; similarly we have $f \geq 0$.
\end{proof}

\subsection{Basic probability}

\begin{lemma}\label{lemma:log-change-of-measure}
Let $\mu, \nu$ be distributions on set $\cX$ with positive density, and let $f: \cX\to[0,1]$ be a function. Then
\[\EE_\nu[f] - \EE_\mu[f] \leq \EE_\mu[f\log(\nu/\mu)] + \Dkl{\mu}{\nu}.\]
\end{lemma}

\begin{proof}
Observe that the function $z \mapsto -\log(z)+z-1$ is nonnegative on $(0,\infty)$. Thus, for all $x \in \cX$,
\[(1-f(x))(-\log(\nu(x)/\mu(x)) + \nu(x)/\mu(x) - 1) \geq 0.\]
Taking expectation over $x \sim \mu$, we get
\[\EE_\nu[f] - \EE_\mu[f] \leq \EE_\mu[f\log(\nu/\mu)] + \EE_\mu[\log(\mu/\nu) + \nu/\mu - 1] = \EE_\mu[f\log(\nu/\mu)] + \Dkl{\mu}{\nu}\]
as claimed.
\end{proof}

The following lemma is a straightforward calculation via convexity of KL-divergence.

\begin{lemma}\label{lemma:kl-ub}
Let $p$ be a distribution on $\RR^d$, supported in a Euclidean ball of radius $R \geq 1$. Then the convolution $p \ast \normal(0, \sigma^2 I_d)$ satisfies
\[\Dkl{p\ast\normal(0,\sigma^2 I_d)}{\normal(0,I_d)} \leq R^2 + d(\sigma^2 + \log(1/\sigma)).\]
\end{lemma}

\subsection{Ornstein-Uhlenbeck}

\begin{lemma}\label{lemma:tv-continuity-convergence}
Let $p,q$ be distributions on $\RR^d$ that admit a continuous density. Then the map $t \mapsto \TV(p_t,q_t)$ is continuous in $t$, and $\TV(p_t,q_t) \to 0$ as $t \to \infty$.
\end{lemma}

\begin{proof}
Since total variation distance is invariant to invertible transformation, it suffices to prove the analogous statements for the laws
\[p_t := \Law(X + \sqrt{t}Z)\]
\[q_t := \Law(Y + \sqrt{t}Z)\]
where $X \sim p$, $Y \sim q$, and $Z \sim \normal(0,I_d)$. 

To prove continuity it suffices (by the triangle inequality for total variation) to show that $\TV(p_s, p_t) \to 0$ as $t \to s$. This is a standard property of the diffusion operator \cite[Theorem 1.1]{bogachev2018ornstein}.

To show that $\TV(p_t, q_t) \to 0$ as $t \to \infty$, fix some $\epsilon>0$. Let $\mu,\nu$ be compactly supported distributions with $\TV(\mu,p) \leq \epsilon/3$ and $\TV(\nu,q) \leq \epsilon/3$. Say that $\mu,\nu$ are supported on the ball $B_2(R)$. Since $\sup_{a,b\in B_2(R)} \TV(\normal(a,tI_d), \normal(b,tI_d)) \to 0$ as $t \to \infty$, it follows from a coupling argument that $\TV(\mu_t,\nu_t) \to 0$ as well. For any fixed $t$, by the data processing inequality and triangle inequality,
\[\TV(p_t, q_t) \leq \TV(p, \mu) + \TV(\mu_t, \nu_t) + \TV(\nu, q).\]
Thus, $\TV(p_t,q_t) < \epsilon$ for sufficiently large $t$. 
\end{proof}

The following standard lemma follows from subadditivity of the Poincar\'e constant under convolution \cite{courtade2020bounds}:

\begin{lemma}[Exponential attenuation of Poincar\'e constant]\label{lem:poincare-decay}
Let $p$ be a distribution on $\RR^d$. For any $t \geq 0$, it holds that \[C_P(p_t)\leq e^{-2t}C_P(p)+1-e^{-2t}.\]
\end{lemma}

\begin{lemma}\label{lem:bounded-radius-poincare}
Let $p$ be a distribution on $\RR^d$, supported on the Euclidean ball of radius $R \geq 1$. Then $C_P(p_{\tstar}) \leq 2$ where $\tstar := 1+\log(R)$.
\end{lemma}

\begin{proof}
Note that $p_t$ is the convolution $\normal(0, (1-e^{-2t})I_d)$ with a distribution supported on a ball of radius $e^{-t} R$. Applying \cite[Theorem 1.2]{bardet2018functional} gives
\[C_P(p_t) \leq (1-e^{-2t})\exp\left(\frac{4e^{-2t}R^2}{1-e^{-2t}}\right)\]
from which the claim can be deduced.
\end{proof}

\begin{lemma}\label{lemma:lipschitz-score}
Let $p$ be a distribution on $\RR^d$, supported in a Euclidean ball of radius $R \geq 1$. Let $t_0>0$. Then $\grad \log p_t(x)$ is $O(1 + R^2/t_0^2)$-Lipschitz for all $t \geq t_0$.  
\end{lemma}

\begin{proof}
This result follows from a second-order Tweedie formula for the Hessian of $\log p_t$ \cite{zhang2023moment}.
\end{proof}

\subsection{Stochastic calculus}

The following are standard results from stochastic calculus, or (for the below version of Girsanov's theorem) consequences of standard arguments; while we make an attempt to be rigorous about the necessary regularity conditions, a reader interested in these details should consult an authoritative source \cite{shreve1991brownian,Chewi26Book}.

\begin{lemma}[It\^o's lemma]\label{lemma:ito-formula}
Let $(X_t)_{t \in [0,T]}$ be a stochastic process on $\RR^d$ defined by the SDE
\[\dd X_t = b(X_t, t) \, \dd t + \sqrt{2} \dd W_t.\]
Let $g: \RR^d \times [0,T] \to \RR$ be a twice continuously differentiable function. Then
\[g(X_T,T) - g(X_0,0) = \int_0^T (\partial_t g(X_t,t) + \langle b(X_t,t), \grad g(X_t,t)\rangle + \laplace g(X_t,t)) \, \dd t + \sqrt{2}\int_0^T \langle \grad g(X_t, t), \dd W_t\rangle.\]
\end{lemma}

\begin{lemma}[It\^o's isometry; inner product form]\label{lemma:ito-isometry}
Let $(X_t)_{t \in [0,T]}$ be a stochastic process on $\RR^d$ defined by the SDE
\[\dd X_t = b(X_t, t) \, \dd t + \sqrt{2} \dd W_t.\]
Let $U, V: \RR^d \times [0,T] \to \RR^d$ satisfy $\EE\int_0^T \norm{U(X_t,t)}^2 \, \dd t < \infty$ and $ \EE\int_0^T \norm{V(X_t,t)}^2 \, \dd t  < \infty$. Then
\[\EE \left(\int_0^T \langle U(X_t,t), \dd W_t\rangle\right) \left(\int_0^T \langle V(X_t,t), \dd W_t\rangle\right) = \EE\int_0^T \langle U(X_t,t), V(X_t,t)\rangle \,\dd t.\]
\end{lemma}

\begin{lemma}[Feynman-Kac; see e.g. {\cite[Theorem 2.4 \& Remark 2.6]{palczewski2025partial}}]\label{lemma:feynman-kac}
Let $b: \RR^d \times [0,T] \to\RR^d$ be Lipschitz on $\RR^d$, uniformly in $[0,T]$, and jointly continuous. Let $(X_t)_{t \in [0,T]}$ be a stochastic process on $\RR^d$ defined by the SDE
\[\dd X_t = b(X_t, t) \, \dd t + \sqrt{2} \dd W_t.\]
Let $f: \RR^d \to [0,1]$ be a uniformly continuous function, and for $t \in [0,T]$, define
\[f_t(x) := \EE[f(X_T) \mid{} X_t = x].\]
Then $g(x,t) := f_t(x)$ satisfies the following partial differential equation:
\[\partial_t g + \langle b, \grad g\rangle + \laplace g = 0.\]
\end{lemma}

\begin{lemma}[Localized Girsanov's theorem]\label{lemma:girsanov}
Let $b,\wh b: \RR^d \times [0,T] \to\RR^d$ be Lipschitz on $\RR^d$, uniformly in $[0,T]$. Let $(X_t)_{t \in [0,T]}$, $(\wh X_t)_{t \in [0,T]}$ be stochastic processes on $\RR^d$ defined by the SDEs
\[\dd X_t = b(X_t, t) \, \dd t + \sqrt{2} \dd W_t,\]
\[\dd \wh X_t = \wh b(\wh X_t, t) \, \dd t + \sqrt{2} \dd \wh W_t,\]
with the same initial law. Let $P, \wh P$ denote the respective path laws.

Suppose that
\[\EE \int_0^T \norm{\wh b(X_t, t) - b(X_t, t)}^2 \,\dd t < \infty.\]
Then,
\[\log \frac{\wh P(X)}{P(X)} = \frac{1}{\sqrt{2}}\int_0^{T} \langle \wh b(X_t, t) - b(X_t, t), \dd W_t\rangle - \frac{1}{4}\int_0^{T} \norm{\wh b(X_t, t) - b(X_t, t)}^2 \dd t\]
holds $P$-almost surely.
\end{lemma}

\begin{proof}
This result largely follows from the limiting approximation argument given in the proof of \cite[Theorem 9]{chen2023sampling}; they define a sequence of stopping times $T^n$, and a sequence of stochastic processes $\wh X^n = (\wh X_t^n)_t$ defined by SDEs that switch to drift $b$ after the stopping time $T^n$, and apply Girsanov's theorem to each process. They then use a limiting argument to transfer to the process $\wh X = (\wh X_t)_t$.

The difference between their application and ours is that they only care about computing $\Dkl{P}{\wh P}$, whereas we require a pointwise expression for the log density-ratio. As a result, to make the limiting argument work, we require an explicit choice of stopping times, and we must use the uniform Lipschitz property of $b, \wh b$. 

Let $\CC$ denote the canonical path space; a path $\omega\in\CC$ is a map $\omega: [0,T] \to \RR^d$. Let $A_t(\omega)$ denote the accumulated $L^2$ error on path $\omega$ up to time $t$, i.e.
\[A_t(\omega) := \int_0^t \norm{b(\omega(s),s) - \wh b(\omega(s),s)}^2 \, \dd s.\]
Since $\EE_P A_T < \infty$, there is some $\omega^\star\in\CC$ with $A_T(\omega^\star)<\infty$, but then the uniform Lipschitz bound implies that $A_T(\omega)<\infty$ for all $\omega\in\CC$.

\paragraph{Stopping time construction and localization.} For every $n>0$, we define the stopping time
\[T_n(\omega) := \inf\{t\in[0,T]: A_t(\omega) \geq n\} \land T.\]
Define the path law $\wh P^n$ by
\begin{equation} \log \frac{\dd \wh P^n(X)}{\dd P(X)} := H^n(X) := \frac{1}{\sqrt{2}}\int_0^{T_n} \langle \wh b(X_s,s)- b(X_s,s), \dd W_s\rangle - \frac{1}{4}A_{T_n}(X).\label{eq:localized-girsanov}\end{equation}
By the same argument as in \cite{chen2023sampling}, Girsanov's theorem implies that, under $\wh P^n$,
\[\dd X_t = (\indic[t\leq T_n]\wh b(X_t,t) + \indic[t>T_n]b(X_t,t))\dd t + \sqrt{2} \dd W_t^n\]
where $W_t^n$ can be defined (in terms of $W_t$ and $X$) so that it is a Brownian motion under $\wh P^n$. We now must use this SDE to argue that both sides of \eqref{eq:localized-girsanov} converge $P$-almost surely to the corresponding ``unstopped'' quantities.

\paragraph{Limiting approximation: RHS.} First we argue that $H^n \to H$ holds $P$-almost surely, where
\[H := \frac{1}{\sqrt{2}}\int_0^{T} \langle \wh b(X_s,s)-b(X_s,s), \dd W_s\rangle - \frac{1}{4}A_{T}(X).\]
Indeed, if $T_n = T$ then $H^n = H$, so it suffices to argue that, $P$-almost surely over $\omega \in \CC$, it holds that $T_n(\omega)=T$ for sufficiently large $n$. But this is immediate from the above fact that $A_T(\omega)<\infty$ for all $\omega$.

\paragraph{Limiting approximation: LHS} Next we argue that $\wh P^n/P \to \wh P/P$. Consider the process $Y^n = (Y_t^n)_t$ defined on probability space $(\CC, \wh P^n)$ by the following SDE:
\[\dd Y^n_t = \wh b(Y_t^n, t) \dd t + \sqrt{2} \dd W_t^n, \qquad Y_0^n = X_0.\]
Note that $Y^n$ has law $\wh P$. But we know that $X$ and $Y^n$ coincide on $[0, T_n(X)]$ (formally, this uses pathwise uniqueness of the above SDE) almost surely under $\wh P^n$. Thus, by the coupling bound on total variation,
\[\TV(\wh P, \wh P^n) \leq \wh P^n(Y^n \neq X) \leq \wh P^n(T_n(X) < T).\]
If $T_n(X) < T$ then $A_{T_n(X)}(X) \geq n$, and so $A_{T_n(X)}(Y^n) \geq n$ (since $X$ and $Y^n$ agree up to $T_n(X)$), which means that $A_T(Y^n) \geq n$. Thus,
\[\wh P^n(T_n(X)<T) \leq \wh P^n(A_T(Y^n) \geq n) = \wh P(A_T \geq n) \to 0\]
where the convergence holds since we showed that $A_T(\omega)$ is finite for all $\omega$. Thus, $\TV(\wh P^n, \wh P) \to 0$ as $n \to \infty$.

Since $\wh P^n \ll P$ for all $n$ by construction and $\TV(\wh P^n,\wh P) \to 0$, it follows that $\wh P \ll P$ . Thus, we may define $I := \wh P/P$. Rewriting the statement $\TV(\wh P^n,\wh P) \to 0$, we have
\[\lim_{n\to \infty} \EE_P|\exp(H^n) - I| = 0.\]
By itself this does not imply almost sure convergence, but we proved above that $H^n \to H$ holds $P$-almost surely. Thus, Fatou's lemma gives $\EE_P|\exp(H)-I| = 0$, and hence $\exp(H) = I$ almost surely. This completes the proof.
\end{proof}

\end{document}